\documentclass[runningheads]{llncs}
\usepackage{graphicx}
\usepackage{tikz}
\usepackage{comment}
\usepackage{amsmath,amssymb} % define this before the line numbering.
\usepackage{color}

\usepackage[accsupp]{axessibility}  % Improves PDF readability for those with disabilities.

\definecolor{azure}{rgb}{0.3, 0.5, 0.9}

\newcommand{\eg}{e.g.}
\newcommand{\ie}{i.e.}
\usepackage[pagebackref=true,breaklinks=true,colorlinks,bookmarks=false]{hyperref}
\usepackage{makecell}
\usepackage{boldline}
\usepackage{algorithm, algorithmic}
\usepackage{multirow}
\usepackage{booktabs}
\usepackage{subcaption}
\usepackage{wrapfig}
\usepackage{enumitem}
\usepackage[capitalize]{cleveref}

\usepackage{lipsum}
\newcommand\blfootnote[1]{%
\begingroup
\renewcommand\thefootnote{}\footnote{#1}%
\addtocounter{footnote}{-1}%
\endgroup
}

\crefname{section}{Sec.}{Secs.}
\Crefname{section}{Section}{Sections}
\Crefname{table}{Table}{Tables}
\crefname{table}{Tab.}{Tabs.}
\crefname{table}{Tab.}{Tabs.}

\begin{document}
% \linenumbers
\pagestyle{headings}
\mainmatter
\def\ECCVSubNumber{7845}  % Insert your submission number here

\title{MVDG: A Unified Multi-view Framework for Domain Generalization} % Replace with your title
 
%\begin{comment} 
\titlerunning{MVDG: A Unified Multi-view Framework for Domain Generalization}
\authorrunning{Jian Zhang\and Lei Qi \and Yinghuan Shi \and Yang Gao}

\author{Jian Zhang\inst{1,2} \and Lei Qi\inst{3}$^{,*}$ \and Yinghuan Shi\inst{1,2}$^{,*}$ \and Yang Gao\inst{1,2}}
\institute{State Key Laboratory for Novel Software Technology, Nanjing University, China. \and
National Institute of Healthcare Data Science, Nanjing University, China.\and
School of Computer Science and Engineering, Southeast University, China.  
\blfootnote{* Corresponding Authors: Yinghuan Shi (syh@nju.edu.cn) and Lei Qi (qilei@seu.edu.cn).}
}
%\end{comment}
%******************

\begin{comment}
\titlerunning{Abbreviated paper title}
\author{First Author\inst{1}\orcidID{0000-1111-2222-3333} \and
  Second Author\inst{2,3}\orcidID{1111-2222-3333-4444} \and
  Third Author\inst{3}\orcidID{2222--3333-4444-5555}}
%
\authorrunning{F. Author et al.}
\institute{Princeton University, Princeton NJ 08544, USA \and
  Springer Heidelberg, Tiergartenstr. 17, 69121 Heidelberg, Germany
  \email{lncs@springer.com}\\
  \url{http://www.springer.com/gp/computer-science/lncs} \and
  ABC Institute, Rupert-Karls-University Heidelberg, Heidelberg, Germany\\
  \email{\{abc,lncs\}@uni-heidelberg.de}}
\end{comment}
%******************
\maketitle

%%%%%%%%% ABSTRACT  70-150 words limitation?
\begin{abstract}
  
  To generalize the model trained in source domains to unseen target domains, domain generalization (DG) has recently attracted lots of attention. Since target domains can not be involved in training, overfitting source domains is inevitable. As a popular regularization technique, the meta-learning training scheme has shown its ability to resist overfitting. However, in the training stage, current meta-learning-based methods utilize only one task along a single optimization trajectory, which might produce a biased and noisy optimization direction. Beyond the training stage, overfitting could also cause unstable prediction in the test stage. In this paper, we propose a novel multi-view DG framework to effectively reduce the overfitting in both the training and test stage. {Specifically, in the training stage, we develop a multi-view regularized meta-learning algorithm that employs multiple optimization trajectories to produce a suitable optimization direction for model updating. We also theoretically show that the generalization bound could be reduced by increasing the number of tasks in each trajectory.} In the test stage, we utilize multiple augmented images to yield a multi-view prediction to alleviate unstable prediction, which significantly promotes model reliability. Extensive experiments on three benchmark datasets validate that our method can find a flat minimum to enhance generalization and outperform several state-of-the-art approaches. The code is available at \url{https://github.com/koncle/MVDG}. 
  
\end{abstract}
%%%%%%%%% BODY TEXT

\section{Introduction}

\begin{figure}[t]
  \begin{center}
    \includegraphics[width=0.6\linewidth]{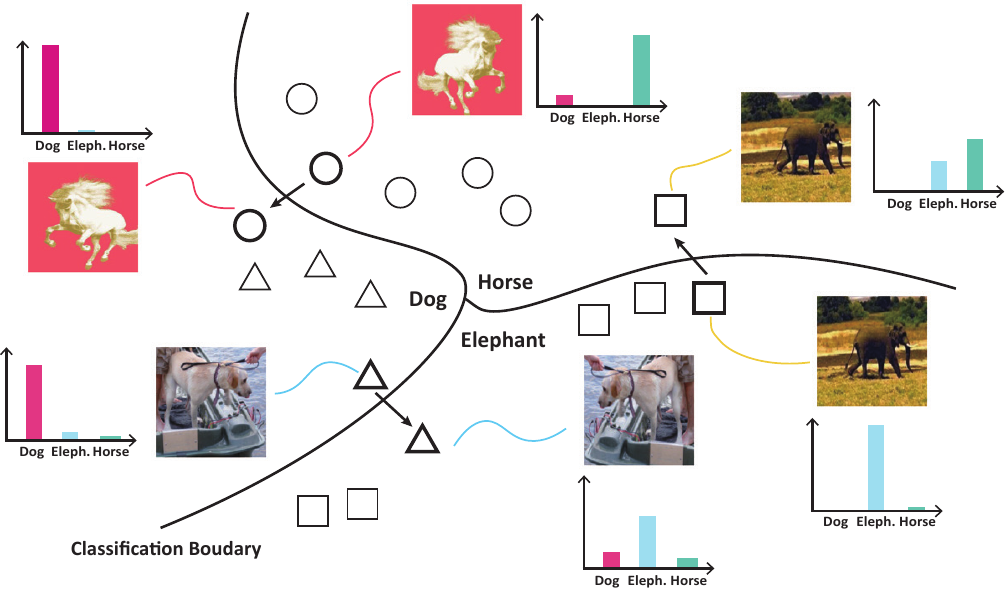}
  \end{center}
  %\vspace{-10pt} 
  \caption{The examples for the variation of the prediction probability when the test images are slightly perturbed in the unseen domain on the PACS dataset.}
  \label{fig:tta_conf}
  %%\vspace{-0.4cm}
  %\vspace{-15pt}
\end{figure}

Traditional supervised learning assumes that training and test data are from the same distribution. However, this conventional assumption is not always satisfied in the real world when domain shifts exist between training and test data. Recently, learning a robust and effective model against domain shift has raised considerable attention~\cite{ben2007analysis,kouw2019review}. As one of the most representative learning paradigms under domain shift, unsupervised domain adaption (UDA) aims to tackle the adaptation from the labeled source domain to the unlabeled target domain under domain shift.
Despite the great success of current UDA models~\cite{na2021fixbi,melas2021pixmatch,yue2021transporting}, when deploying the previously trained UDA model to other unseen domains, we should re-train the model by incorporating the newly collected data from the unseen domain. This re-training process not only increases extra space/time costs but also violates privacy policy in some cases (\eg, clinical data), rendering these UDA methods not applicable to some real tasks.

The dilemma above motivates us to focus on a more applicable yet challenging setting, namely domain generalization (DG) \cite{muandet2013domain}. In DG, by only learning the related knowledge from existing source domains, the trained model is required to be directly applied to previously unseen domains without any re-training procedure.
To guarantee the model's efficacy on the unseen target domain, previous DG methods~\cite{muandet2013domain,rahman2019correlation,li2018domain} intend to reduce the domain-specific influence from source domains via learning domain-invariant representations.

{Well fitting source domains is easy, whereas well generalizing to an unseen target domain is hard. Previous methods inevitably suffer from overfitting when concentrating only on fitting source domains.}
Therefore, the meta-learning-based methods~\cite{li2018learning,dou2019domain} have arisen as one of the most popular methods to resist overfitting during training, which simulates the domain shift episodically to perform regularization. However, these methods train their model with a single task at each iteration, which could cause a biased and noisy optimization direction.

% When a model overfits source domains, it possibly makes a biased prediction for \textcolor{red}{the perturbed (not mentioned)} image. 
Besides, after investigating the predictions of the trained model during the test stage, we notice that overfitting also results in unstable predictions. We conduct an experiment by perturbing~(\eg, random crop and flip) the test images. As shown in \cref{fig:tta_conf}, their predictions usually changed after being perturbed.
It is because the feature representations of unseen images learned by the overfitted model are more likely to lie near the decision boundary. These representations are easily perturbed across the boundary, thus, producing different predictions. {This phenomenon is more challenging in DG due to the domain discrepancy.}

As aforementioned, the overfitting problem not only appears in the training stage but also largely influences the following test procedure. To fight against overfitting, we innovatively propose a multi-view framework to deal with the inferior generalization ability and unstable prediction. Specifically, in the training stage, we design a multi-view regularized meta-learning algorithm that can regularize the training process in a simple yet effective way. This algorithm contains two steps. The first step is to guide a model to pursue a suitable optimization direction via exploiting multi-view information. {Unlike most previous methods (e.g., MLDG~\cite{li2018learning}) that train the model using only one task along a single optimization trajectory (\ie, the training path from the current model to another model) with limited single-view information}, we propose to train the model using multiple trajectories with different sampled tasks to find a more accurate direction with integrated multi-view information. In the second step, we update the model with the learned direction. We present a theoretical analysis that the number of tasks is also critical for generalization ability. {Besides, we empirically verify that integrating multiple trajectory information can help our method find a flat minimum for promising generalization.}

In the test stage, we propose to deal with the unstable prediction caused by the overfitted model using multi-view prediction. We argue that current test images with a single view cannot prevent unstable prediction. Nevertheless, different augmentations applied to the test image can bring abundant information from different views. Thus, if using the image pre-processing perturbations in the test procedure (e.g., the cropping operation), we can obtain multi-view information for a single image.
Therefore, we augment each test image into multiple views during the test stage and ensemble their predictions as the final output. By exploiting the multi-view predictions of a single image, we can eliminate the unreliability of predictions and obtain a more robust and accurate prediction.

{
In summary, we propose a unified multi-view framework to enhance the generalization of the model and stabilize the prediction in both training and test stages. Our contributions can be summarized as follows:
\begin{itemize}[noitemsep,topsep=1pt]
  \item During training, we design an effective multi-view regularized meta-learning scheme to prevent overfitting and find a flat minimum that generalizes better.
  \item We theoretically prove that increasing the number of tasks in the training stage results in a smaller generalization gap and better generalizability.
  \item During the test stage, we introduce a multi-view prediction strategy to boost the reliability of predictions by exploiting multi-view information.
  \item Our method is validated via conducting extensive experiments on multiple DG benchmark datasets and outperforms other state-of-the-art methods.
\end{itemize}
}

\section{Related Work}

\textbf{Domain Generalization.} Domain generalization (DG) has been proposed recently to deal with learning the generalizability to the unseen target domain.
Current DG methods can be roughly classified into three categories: \emph{Domain-invariant feature learning}, \emph{Data augmentation} and \emph{Regularization}.

Domain-invariant feature learning-based methods aim to extract domain-invariant features that are applicable to any other domains. One widely employed technique is adversarial training~\cite{li2018deep,rahman2019correlation,zhao2020domain}, which learns the domain-invariant features by reducing the domain gap between multiple source domains. Instead of directly learning the domain-invariant features, several methods \cite{bousmalis2016domain,piratla2020efficient,chattopadhyay2020learning}  try to disentangle it from domain-specific features.

Data augmentation based methods try to reduce the distance between source and unseen domains via augmenting unseen data. Most of them perform image-level augmentation~\cite{yue2019domain,li2021progressive,xu2020robust,zhou2020deep} that generates images with generative adversarial networks~\cite{goodfellow2014generative} or adaptive instance normalization~\cite{huang2017arbitrary}. Since feature statistics contains style information and easy to manipulate, other methods augment features by modifying its statistics~\cite{zhou2021mixstyle,jeon2021feature,nuriel2020permuted} or injecting generated noise~\cite{li2021simple,shu2020prepare}.

As overfitting hurts the generalization ability of a model, regularization-based methods prevent this dilemma by regularizing the model during training. Several works~\cite{carlucci2019domain,wang2020learning} add auxiliary self-supervised loss to perform regularization. \cite{li2018learning,dou2019domain,zhang2022generalizable} adopt a meta-learning framework to regularize the model by simulating domain shift during training. {Ensemble learning~\cite{cha2021swad,zhou2021domain} also has been employed to regularize training models.} 
Our method also belongs to this category which can better prevent overfitting by exploiting multi-view information.

\textbf{Meta-Learning.} Meta-learning~\cite{thrunlorien} is a long-standing topic that learns how to learn. Recently, MAML~\cite{finn2017model} has been proposed as a simple model-agnostic method to learn the meta-knowledge, attracting lots of attention. Due to the unpredictable large computational cost of second-order derivatives, first-order methods are thus developed~\cite{finn2017model,nichol2018reptile} to reduce the cost. Later, meta-learning is introduced into DG to learn generalizable representation across domains. These methods perform regularization under the meta-learning framework. For example, MLDG~\cite{li2018learning} utilizes the episodic training paradigm and updates the network with simulated meta-train and meta-test data. MetaReg~\cite{balaji2018metareg} explicitly learns regularization weights, and Feature-critic~\cite{li2019feature} designs a feature-critic loss to ensure that the updated network should perform better than the original network. Recent DG methods~\cite{qiao2020learning,liu2020shape} all adopt the episodic training scheme similar to MLDG due to its effectiveness. Although these methods can alleviate overfitting by training on a single task, they may produce biased optimization direction. However, our method can mitigate this problem by learning from multiple tasks and optimization trajectories.

\textbf{Test Time Augmentation.} The test time augmentation~(TTA) is originally proposed in~\cite{krizhevsky2012imagenet}, which integrates the predictions of several augmented images for improving the robustness of the final prediction. {Besides, several methods~\cite{molchanov2020greedy,sypetkowski2020augmentation} try to learn automatic augmentation strategy for TTA. Other methods apply TTA to estimate uncertainty in the model~\cite{ayhan2018test,lee2021test}. }
However, according to our best knowledge, TTA has not been explored in DG, which can alleviate prediction uncertainty via generating multi-view augmented test images.

\section{Our Method}

\subsection{Episodic Training Framework}

Given the data space and label space as $\mathcal{X}$ and $\mathcal{Y}$, respectively, we denote $N$ source domains as $\mathcal{D}_1, \dots, \mathcal{D}_N$, where $\mathcal{D}_i=\{x_k, y_k\}_1^{N_i}$. $N_i$ is the number of samples in the $i$-th source domain $\mathcal{D}_i$. We denote the model parameterized by $\theta$ as $f$. Given input $x$, the model outputs $f(x|\theta)$.
As previously reviewed, meta-learning-based methods usually train the model with an episodic training paradigm.
Similar to the meta-learning based methods~\cite{li2018learning} that split the source domains into meta-train and meta-test sets at each iteration, we leave one domain out as meta-test domain $\mathcal{D}_{te}$ and the remaining domains as meta-train domains $\mathcal{D}_{tr}$. Hereafter, we sample mini-batches from these domains and obtain meta-train $\mathcal{B}_{\mathrm{tr}}$ and meta-test data $\mathcal{B}_{\mathrm{te}}$.  A \textit{{sub-DG} task} is defined as a pair of them $t=(\mathcal{B}_{\mathrm{tr}}, \mathcal{B}_{\mathrm{te}})$. We define the loss on a batch  $\mathcal{B} \in \{\mathcal{B}_{\mathrm{tr}}, \mathcal{B}_{\mathrm{te}}\}$ with parameter $\theta$ as:
$$
  \mathcal{L}(\mathcal{B} | \theta) = \sum_{(x_i, y_i)\in \mathcal{B}} \ell \big(f(x_i|\theta), y_i\big),
$$
where $\ell$ is the traditional cross-entropy loss.

\begin{wrapfigure}{r}{0.38\textwidth}
  \vskip -30pt 
  \centering
  \begin{subfigure}{1\linewidth}
    \includegraphics[width=1\linewidth]{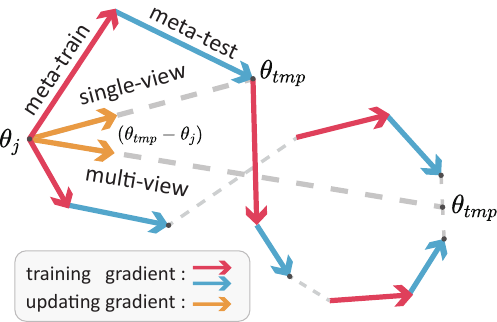}
  \end{subfigure}
  \caption{An illustration of Reptile and our multi-view regularized meta-learning algorithm (best viewed in color).}
  \label{fig:mtml}
  \vskip -20pt
\end{wrapfigure} 

Different from previous meta-learning algorithms (\eg, MLDG~\cite{li2018learning}) that take second-order gradients to update model parameters, Reptile~\cite{nichol2018reptile} is proposed as a first-order algorithm to reduce the computational cost. Therefore, we adopt Reptile version of MLDG. To be specific, given model parameter $\theta_j$ at the $j$-${th}$ iteration, we sample a task and train the model first with $\mathcal{L}(\mathcal{B}_{\mathrm{tr}} | \theta_j)$ and then $\mathcal{L}(\mathcal{B}_{\mathrm{te}} | \theta_j)$. Afterwards, we can obtain a temporarily updated parameter $\theta_{tmp}$ along this optimization trajectory. Finally, we take $(\theta_{tmp}-\theta_j)$ as an optimization direction, \ie, the direction points from the temporary parameter to the original parameter, to update the original parameter $\theta_j$:
 
\begin{equation}
  \label{eq:reptile}
  \theta_{j+1} = \theta_j + \beta (\theta_{tmp} - \theta_j).
\end{equation}  In this way, $\theta_{tmp}$ could exploit a part of current weight space and find a better optimization direction for current sampled tasks.

\subsection{Multi-view Regularized Meta-Learning}
\label{Sec:MVRML}

\begin{wrapfigure}{tr}{0.4\textwidth}
  \vskip -27pt
  \centering

  \begin{subfigure}{0.48\linewidth}
    \includegraphics[width=1\linewidth]{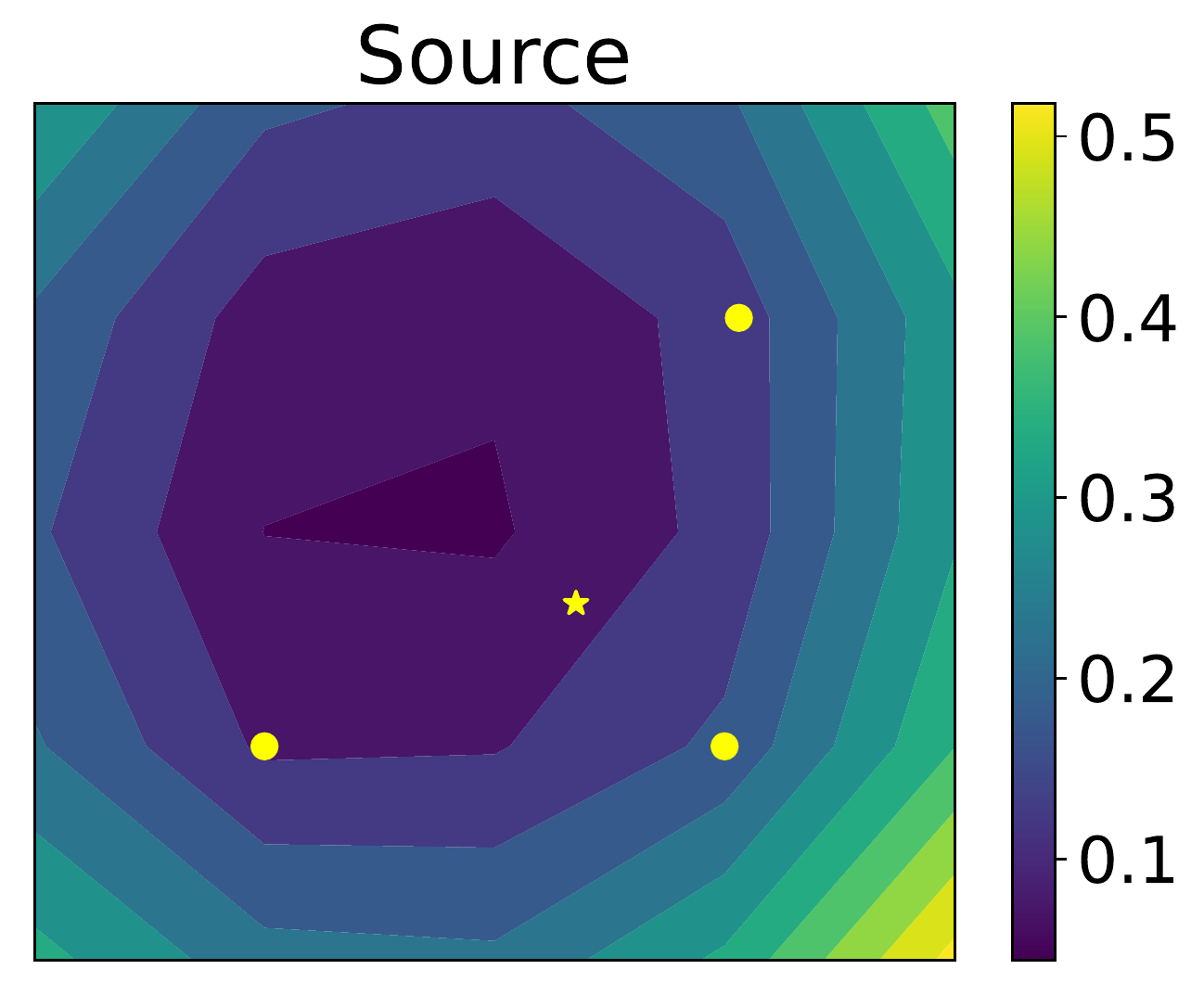}
  \end{subfigure}
  \begin{subfigure}{0.48\linewidth}
    \includegraphics[width=1\linewidth]{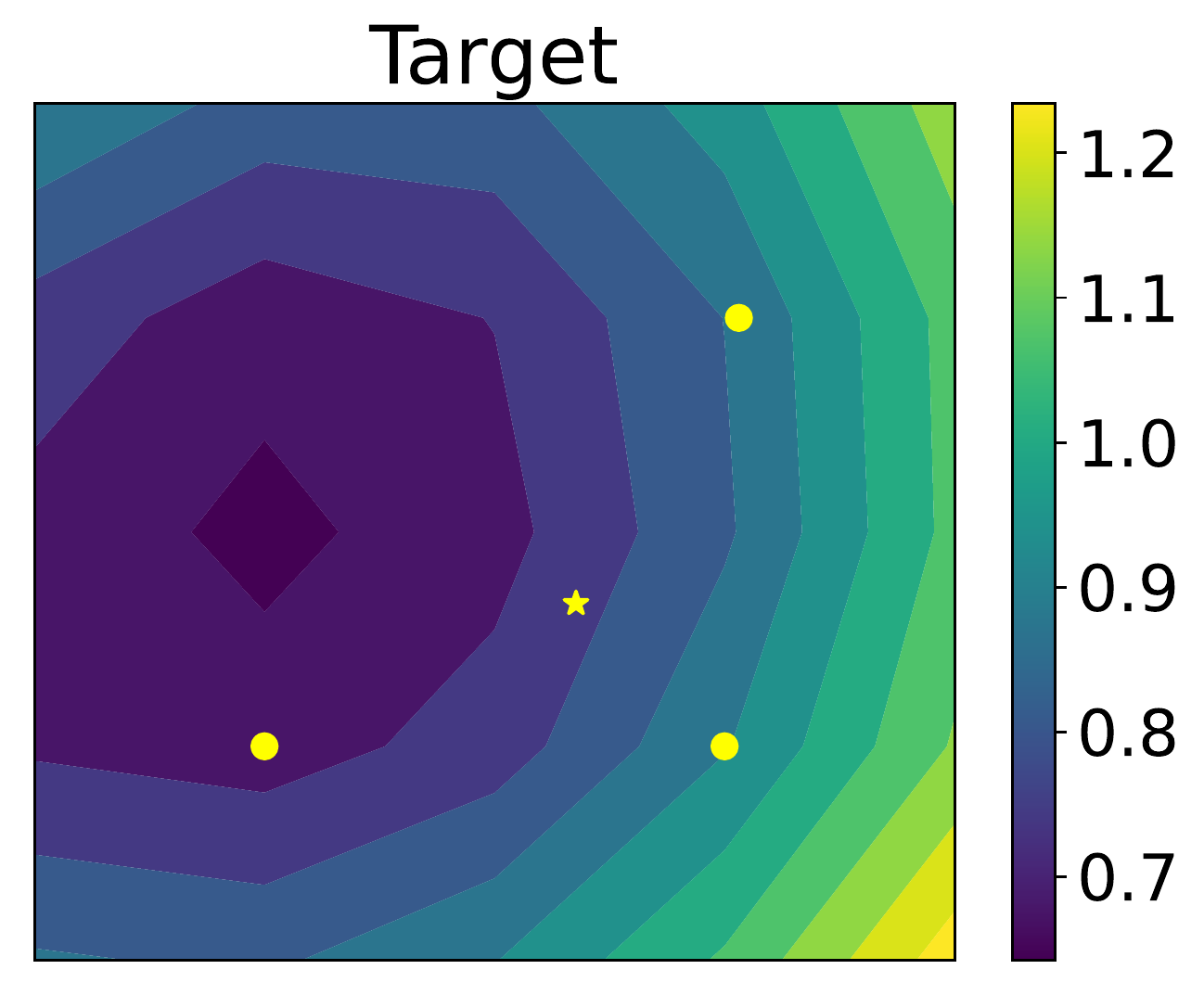}
  \end{subfigure}

  \caption{The loss surface~\cite{garipov2018loss} of three temporarily trained parameters (denoted as dot) starting from $\theta_j$ and optimized along a different trajectory. The star denotes the averaged parameter.}
  \label{fig:flat}
  \vskip -25pt
\end{wrapfigure}
Although Reptile reduces the computational cost, the training scheme above suffers from several problems. {Firstly}, the model is trained along a single optimization trajectory, producing only one temporary parameter $\theta_{tmp}$. It is only a partial view of the weight space for the current parameter $\theta_j$, which is noisy and insufficient to produce a robust optimization direction. 
To better understand this problem, we plot the loss surface of three temporarily trained parameters (denoted as a dot)  in \cref{fig:flat}. Two of the three parameters have higher loss values on both source and target domains. If we only consider one view (\ie, using a single temporary parameter), we may not find a good optimization direction for updating. Recent researchs~\cite{cha2021swad,frankle2020linear,izmailov2018averaging} find that the flat minimum can produce better generalization performance than a sharp minimum, and simple weight 
\noindent averaging can achieve it if these weights share part of optimization trajectory~\cite{frankle2020linear}. Inspired by this conclusion, we also plot the averaged parameter (denoted as a star) and find that it can produce a more robust parameter with lower loss than most parameters on both training and test domains. Therefore, the ensembling of temporary models can produce a better and more stable optimization direction.

{Secondly}, since the model is trained with a single task in each trajectory, it cannot explore the weight space fully and is hard to escape from the local minimum, resulting in the overfitting problem. We have theoretically proved that more sampled tasks can help minimize the domain generalization error and produce better performance in \cref{Sec:Theory}.
 
Aiming to better explore the weight space to obtain a more accurate optimization direction and erase the impact of overfitting, we develop a simple yet effective multi-view regularized meta-learning (MVRML) algorithm by exploiting multi-view information at each iteration. Specifically, to find a robust optimization direction, we obtain $T$ temporary parameters $\{\theta_{tmp}^1, ..., \theta_{tmp}^T\}$ along different optimization trajectories. Different from MLDG, which only samples a single task for the training stage, we train each temporary parameter with $s$ sampled tasks to help it escape from the local minimum. Besides, we sample different tasks in different trajectories to encourage the diversity of temporary models. Learning from these tasks allows to explore different information from weight space with supplementary views. This sampling strategy plays an important role in our method, and we will verify it in \cref{sec:exp_further}. Then we average their weights to obtain a robust parameter: $\theta_{tmp}=\frac{1}{T}\sum_{t=1}^T\theta_{tmp}^t$. Since these models share a part of the optimization trajectory, ensembling their weights can help find parameters located in the flat minimum that generalizes better. The full algorithm is shown in \cref{alg:mt} and illustrated in \cref{fig:mtml}.

\begin{figure}[t]
  \begin{algorithm}[H]
    \caption{Multi-view Regularized Meta-Learning.}
    % \definecolor{babyblue}{rgb}{}
    \definecolor{a}{rgb}{0.19, 0.55, 0.91}
    \definecolor{b}{rgb}{0.4, 0.6, 0.8}
    \renewcommand{\algorithmiccomment}[1]{#1}
    \textbf{Input: }
    source data $\mathcal{D}_{src}$, network parametrized by $\theta$, hyperparameters: inner loop learning rate $\alpha$, outer loop learning rate $\beta$, the number of optimization trajectories $T$ and the number of sampled tasks $s$. \\
    \textbf{Output: } the trained parameter
    \begin{algorithmic}[1]\label{alg:mt}
      \STATE $\theta_0 \leftarrow \theta; j\leftarrow 0$ \textcolor{b}{\texttt{// initialize parameter}}
      \WHILE{not converged}
      \STATE $\theta_j^0 \leftarrow$ Initialized by $\theta_j$
      \FOR[\textcolor{b}{\texttt{// train temporary models}}]{$t\in\{1,...,T\}$} 
      \FOR{$i\in\{1,...,s\}$}
      \STATE $\mathcal{D}_{tr}, \mathcal{D}_{te} \leftarrow$ Random split $\mathcal{D}_{src}$   \textcolor{b}{\texttt{// $\mathcal{D}_{tr} \cup \mathcal{D}_{te} = \mathcal{D}_{src}$, $\mathcal{D}_{tr} \cap \mathcal{D}_{te} = \emptyset$}}
      \STATE Sample mini-batch $\mathcal{B}_{tr}, \mathcal{B}_{te}$ from $\mathcal{D}_{tr}, \mathcal{D}_{te}$
      \STATE $\theta_j^{i-1} \leftarrow \theta_j^{i-1} - \alpha  \triangledown_{\theta}(\mathcal{L}(\mathcal{B}_{\mathrm{tr}}|\theta_j^{i-1})) $  \textcolor{b}{\texttt{// train with meta-train data}}
      \STATE $\theta_j^i \leftarrow \theta_j^{i-1} - \alpha  \triangledown_{\theta}(\mathcal{L}(\mathcal{B}_{\mathrm{te}}|\theta_j^{i-1})) $ \textcolor{b}{\texttt{// train with meta-test data}}
      \ENDFOR
      \STATE $\theta_{tmp}^t \leftarrow \theta_j^i$    \textcolor{b}{\texttt{// assign current temporary parameter}}
      \ENDFOR
      \STATE $\theta_{j+1} \leftarrow \theta_j + \beta (\frac{1}{T}\sum_{t=1}^T\theta_{tmp}^t - \theta_j)$ \textcolor{b}{\texttt{// update the original parameter}}
      \STATE $j\leftarrow j+1$
      \ENDWHILE
      % \STATE Re-estimate BN statistics
    \end{algorithmic}
  \end{algorithm}
  %\vspace{-30pt}
\end{figure}

% It is worth noting that, at the end of the training stage, we re-estimate batch normalization~(BN)~\cite{ioffe2015batch} statistics by forwarding the training set. We notice that the test accuracy is unstable, which is caused by the mismatch between BN statistics and model weights. During training, BN normalizes data with statistics calculated in a batch and keeps a running average of its statistics which is used to normalize test images. However, for MVRML, the second step is not done because we only update model weights while leaving its statistics of BN unchanged, which causes a mismatch. Thus, the performance fluctuates when we apply these mismatched weights and statistics to test images. We mitigate this problem by re-estimating BN statistics at the end of the training procedure.

\textbf{Relation to Reptile.} Our algorithm is similar to the batch version of Reptile~\cite{nichol2018reptile}, but there are three key differences. Firstly, the goal is different. Reptile aims to find a good weight initialization that can be fast adapted to new tasks, while since our classification task is fixed, we are more interested in the generalization problem without adaptation. Secondly, the task sampling strategy is different. Reptile only samples a single task for each trajectory, while we sample multiple different tasks for better generalization. Finally, the training scheme is different. In Reptile, since its goal is to adapt to the current classification task, it trains the model on this task iteratively. However, it is easy to overfit this single task in DG since the performance of the training model is already good in the source domains. Differently, we train different tasks in each trajectory to find a more generalizable parameter and prevent the overfitting problem.

\textbf{Relation to Ensemble Learning.} There are two ensemble steps in our method. First, at each iteration, we ensemble several temporary model parameters to find a robust optimization direction. Second, if we change the formulation of \cref{eq:reptile} as $\theta_{i+1} = (1 - \beta) \theta_{i} + \beta \theta_{\mathrm{tmp}},$ we can obtain an ensemble learning algorithm that combines the weights of current model and temporary model. Therefore, this training paradigm implicitly ensembles models in the weight space and can lead to a more robust model~\cite{izmailov2018averaging}.

\subsection{Theoretical Insight}

\label{Sec:Theory}
{Traditional meta-learning methods train the model with only one task, which could suffer from the overfitting problem. We theoretically prove that increasing the number of tasks can improve the generalizability in DG.} 
We denote the source domain as $\mathcal{S}=\{\mathcal{D}_1, ...\mathcal{D}_N\}$ and target domain as $\mathcal{T}=\mathcal{D}_{N+1}$. A task is defined as $t=(\mathcal{B}_{tr},\mathcal{B}_{te})$, which is obtained by sampling from $\mathcal{S}$. At each iteration, a sequence of sampled tasks along a single trajectory is defined as $\mathbf{T}=\{t_0, \dots, t_m\}$ with a size of $m$. The {training set of task sequences} is defined as $\mathbb{T}=\{\mathbf{T}_0, \mathbf{T}_1, \dots, \mathbf{T}_n\}$ with a size of $n$. 
A training algorithm $\mathbb{A}$ trained with $\mathbb{T}$ or $\mathbf{T}$ is denoted as $\theta=\mathbb{A}(\mathbb{T})$ or $\theta=\mathbb{A}(\mathbf{T})$. We define the expected risk as $\mathcal{E}_{\mathcal{P}}(\theta)=\mathbb{E}_{(x_i, y_i)\sim \mathcal{P}} \ell (f(x_i|\theta), y_i)$. With a little abuse of notation, we define the loss with respect to $\mathbf{T}$  as:
\begin{equation}
  \mathcal{L}(\mathbf{T};\theta)=\frac{1}{m}\sum_{(\mathcal{B}_{tr}, \mathcal{B}_{te})\in \mathbf{T}} \frac{1}{2}(\mathcal{L}(\mathcal{B}_{tr};\theta)+\mathcal{L}(\mathcal{B}_{te};\theta)),
\end{equation}
and the loss with respect to $\mathbb{T}$  as 
 $\mathcal{L}({\mathbb{T}};\theta)=\frac{1}{n}\sum_{\mathbf{T}\in \mathbb{T}} \mathcal{L}(\mathbf{T};\theta).$ \\ 

\noindent \textbf{Theorem 1.} {Assume that algorithm $\mathbb{A}$ satisfies $\beta_1$-uniform stability}~\cite{bousquet2002stability} {with respect to $\mathcal{L}(\mathbb{T};\mathbf{A}(\mathbb{T}))$ and $\beta_2$-uniform stability with respect to $\mathcal{L}(\mathbf{T};\mathbf{A}(\mathbf{T}))$. The following bound holds with probability at least $1-\delta$}:

\begin{equation}
  \mathcal{E}_{\mathcal{T}}(\theta)  \le \hat{\mathcal{E}}_{\mathcal{S}}(\theta) + \frac{1}{2}\sup_{\mathcal{D}_i \in\mathcal{S}}\textbf{Div}(\mathcal{D}_i, \mathcal{T}) + 2\beta_1 + (4n\beta_1+M)\sqrt{\frac{\ln \frac{1}{\delta}}{2n}} + 2\beta_2,
\end{equation} 
where  $M$ is a bound of loss function $\ell$ and $\textbf{Div}$ is KL divergence. $\beta_1$ and $\beta_2$ are functions of the number of task sequences $n$ and the number of tasks $m$ in each task samples. When $\beta_1=o(1/n^a), a\ge1/2$ and $\beta_2=o(1/m^b),b\ge0$, this bound becomes non-trivial. Proof is in \textbf{Supplementary Material}.

This bound contains three terms: (1) the empirical error estimated in the source domain; (2) the distance between the source and target domain; (3) the confidence bound related to the number of task sequences $n$ and the number of tasks $m$ in each sequence. Traditional meta-learning methods in DG train with a large number of task sequences (\ie, $n$). However, the number of tasks in the sequence is only one (\ie, $m$). \textit{By increasing the number of sampled tasks in each sequence, we can obtain a lower error bound. In this case, we could expect a better generalization ability.}

\subsection{Multi-view Prediction}

% Although we can alleviate the inferior generalization result caused by the overfitting problem in the training stage, it still exists and will cause prediction instability and performance degradation in the test stage.

As our model is trained in source domains, the feature representations of learned data are well clustered. However, when unseen images come, they are more likely to be near the decision boundary because of the overfitting and domain discrepancy, leading to unstable feature representations. When we apply small perturbations to the test images, their feature representations will be pushed across the boundary, as shown in \cref{fig:tta_conf}.

However, current test images only present a single view (\ie, the original image) with limited information. As a result, it cannot completely prevent the unstable prediction caused by overfitting. Besides, we argue that different views of a single image could bring in more information than a single view.  Therefore, instead of only using a single view to conduct the test, we propose to perform multi-view prediction~(MVP). By performing multi-view predictions, we can integrate complementary information from these views and obtain a more robust and reliable prediction. Assuming we have an image $x$ to be tested, we can generate different views of this image with some weak stochastic transformations $\mathrm{T}(\cdot)$. Then the image prediction $p$ is obtained by:
$$
  p = \mathrm{softmax}\big(\frac{1}{m}\sum_{i=1}^{m} f(\mathrm{T}(x)|\theta)\big),
$$
where $m$ is the number of views for a test image. We only apply the weak transformations (\eg, random flip) for MVP because we find that the strong augmentations (\eg, the color jittering) make the augmented images drift off the manifold of the original images, resulting in unsatisfactory prediction accuracy, {which will be shown in \textbf{Supplementary Material}}. 

{Note that the improvement brought by MVP does not mean the learned model has poor generalization capability on the simple transformations since our method without MVP has superior performance compared to other methods, and MVP can also improve other SOTA models. Besides, most predictions can not be changed if a model is robust. We will verify these claims in \cref*{sec:exp_further}.}

\section{Experiments}
\label{sec:exp}

% We now report both quantitative and qualitative results of our evaluation. Specifically, we first describe the details of datasets and implementation. Then, we extensively compare our method with state-of-the-art methods. Moreover, we conduct the ablation study to confirm the effectiveness of each module used in our framework. Lastly, we analyze the properties of our method systematically. {More experiments can be found in \textbf{Appendix}.}
 
% %\vspace*{-10pt}
% \subsection{Datasets}

% %\vspace*{-5pt}
We describe the details of datasets and implementation details as follows:

\textbf{Datasets.} To evaluate the performance of our method, we consider three popularly used domain generalization datasets: PACS, VLCS and OfficeHome. PACS~\cite{li2017deeper} contains 9,991 images with 7 classes and 4 domains, and there is a large distribution discrepancy across domains. VLCS~\cite{torralba2011unbiased} consists of 10,729 images, including 5 classes and 4 domains with a small domain gap. 
OfficeHome~\cite{venkateswara2017deep} contains 15,500 images, covering 4 domains and 65 categories which are significantly larger than PACS and VLCS. 

\textbf{Implementation details.} We choose ResNet-18 and Resnet-50~\cite{he2016deep} pretrained on ImageNet~\cite{deng2009imagenet} as our backbone, the same as previous methods \cite{dou2019domain}. All images are resized to 224$\times$224, and the batch size is set to 64. The data augmentation consists of random resize and crop with an interval of $[0.8, 1]$, random horizontal flip, and random color jittering with a ratio of $0.4$.  The model is trained for 30 epochs. We use SGD as our outer loop optimizer and Adam as the inner loop optimizer, both with a weight decay of $5\times 10^{-4}$. Their initial learning rates are $0.05$ and $0.001$ for the first 24 epochs, respectively, and they are reduced to $5\times 10^{-3}$ and $1\times10^{-4}$ for the last 6 epochs. The $\beta_1$ and $\beta_2$ are 0.9 and 0.999 for Adam optimizer, respectively. The number of optimization trajectories and sampled tasks are both 3. For multi-view prediction, we only apply weak augmentations, \ie, the random resized crop with an interval of $[0.8, 1]$ and random horizontal flip. The augmentation number $t$ is set to 32. If not specially mentioned, we adopt this implementation as default.

We adopt the leave-one-out~\cite{li2017deeper} experimental protocol that leaves one domain as an unseen domain and other domains as source domains. We conduct all experiments three times and average the results. Following the way in~\cite{li2017deeper}, we select the best model on the validation set of source domains. DeepAll indicates that the model is trained without any other domain generalization modules.

\begin{table}[t]
  \footnotesize
  \caption{Domain generalization accuracy~(\%) on PACS dataset with ResNet-18 (left) and ResNet-50 (right) backbone. The best performance is marked as \textbf{bold}.}
  \begin{center}
    \begin{minipage}{.48\columnwidth}
      \resizebox{1\columnwidth}{!}{
        \begin{tabular}{l |  c c c c | c}
          \toprule
          ~\textbf{Method}~                & \textbf{A}     & \textbf{C}     & \textbf{P}     & \textbf{S}     & \textbf{Avg.}  \\
          \midrule
          DeepAll                          & 78.40          & 75.76          & \textbf{96.49} & 66.21          & 79.21          \\
          MLDG~\cite{li2018learning}       & 79.50          & 77.30          & 94.30          & 71.50          & 80.70          \\
          MASF~\cite{dou2019domain}        & 80.29          & 77.17          & 94.99          & 71.69          & 81.04          \\
          MetaReg~\cite{balaji2018metareg} & 83.70          & 77.20          & 95.50          & 70.30          & 81.70          \\
          VDN~\cite{wang2021variational}   & 82.60          & 78.50          & 94.00          & 82.70          & 84.50          \\
          FACT~\cite{xu2021fourier}        & 85.37          & 78.38          & 95.15          & 79.15          & 84.51          \\
          RSC~\cite{huang2020self}         & 83.43          & 80.31          & 95.99          & 80.85          & 85.15          \\
          FSDCL~\cite{jeon2021feature}     & 85.30          & \textbf{81.31} & 95.63          & 81.19          & 85.86          \\
          \midrule
          MVDG (Ours)                             & \textbf{85.62} & 79.98          & 95.54          & \textbf{85.08} & \textbf{86.56} \\
          \bottomrule
        \end{tabular}
      }
    \end{minipage}
    % \hspace{10pt} 
    \begin{minipage}{.49\columnwidth}
      \resizebox{1\columnwidth}{!}{
        \begin{tabular}{l | c c c c | c}
          \toprule
          ~\textbf{Method}~                & \textbf{A}     & \textbf{C}     & \textbf{P}     & \textbf{S}     & \textbf{Avg.}  \\
          \midrule
          DeepAll                          & 86.72          & 76.25          & \textbf{98.22} & 76.31          & 84.37          \\
          MASF~\cite{dou2019domain}        & 82.89          & 80.49          & 95.01          & 72.29          & 82.67          \\
          MetaReg~\cite{balaji2018metareg} & 82.57          & 79.20          & 97.60          & 70.30          & 83.58          \\
          MatchDG~\cite{mahajan2021domain} & 85.61          & 82.12          & 78.76          & 97.94          & 86.11          \\
          DSON~\cite{seo2019learning}      & 87.04          & 80.62          & 95.99          & 82.90          & 86.64          \\
          RSC~\cite{huang2020self}         & 87.89          & 82.16          & 97.92          & 83.35          & 87.83          \\
          FSDCL~\cite{jeon2021feature}     & 88.48          & 83.83          & 96.59          & 82.92          & 87.96          \\
          SWAD~\cite{cha2021swad}          & 89.30          & 83.40          & 97.30          & 82.50          & 88.10          \\
          \midrule
          MVDG (Ours)                             & \textbf{89.31} & \textbf{84.22} & 97.43          & \textbf{86.36} & \textbf{89.33} \\
          \bottomrule
        \end{tabular}
        \label{tab:pacs}
      }
    \end{minipage}

  \end{center}
  %\vspace{-20pt}
\end{table}

\subsection{Comparison with State-of-the-art Methods}

We evaluate our method (namely \textbf{MVDG}) to different kinds of recent state-of-the-art DG methods on several benchmarks to demonstrate its effectiveness.

\begin{table}[t]
  \footnotesize
  \caption{Domain generalization accuracy~(\%) on VLCS and OfficeHome datasets. The best performance is marked as \textbf{bold}.}
  \begin{center}
    \begin{minipage}{.49\columnwidth}
      \resizebox{1\columnwidth}{!}{
        \begin{tabular}{l | c c c c | c}
          \toprule
          ~\textbf{Method}~               & \textbf{C}     & \textbf{L}     & \textbf{P}     & \textbf{S}     & \textbf{Avg.}  \\
          \midrule
          DeepAll                         & 96.98          & 62.00          & 73.83          & 68.66          & 75.37          \\
          JiGen~\cite{carlucci2019domain} & 96.17          & 62.06          & 70.93          & 71.40          & 75.14          \\
          MMLD~\cite{matsuura2020domain}  & 97.01          & 62.20          & 73.01          & \textbf{72.49}          & 76.18          \\
          RSC~\cite{huang2020self}        & 96.21          & 62.51          & 73.81          & 72.10          & 76.16          \\
          % StableNet~\cite{zhang2021deep}  & 96.67          & \textbf{65.36} & 73.59          & \textbf{74.97} & \textbf{77.65} \\
          \midrule
          MVDG (Ours)                            & \textbf{98.40} & \textbf{63.79}          & \textbf{75.26} & 71.05          & \textbf{77.13}          \\
          \bottomrule
        \end{tabular}
      }
      \label{tab:vlcs}
    \end{minipage}
    \begin{minipage}{.49\columnwidth}
      \resizebox{1\columnwidth}{!}{
        \begin{tabular}{l | c c c c | c}
          \toprule
          ~\textbf{Method}~            & \textbf{A}     & \textbf{C}     & \textbf{P}     & \textbf{R}     & \textbf{Avg.}  \\
          \midrule
          DeepAll                      & 58.65          & 50.35          & 73.74          & 75.67          & 64.60          \\
          % JiGen~\cite{carlucci2019domain} & 53.04          & 47.51          & 71.47          & 72.29          & 61.20          \\
          % DSON~\cite{seo2019learning}     & 59.37          & 45.70          & 71.84          & 74.68          & 62.90          \\
          RSC~\cite{huang2020self}     & 58.42          & 47.90          & 71.63          & 74.54          & 63.12          \\
          % CrossGrad~\cite{shankar2018generalizing}  & 58.40          & 49.40          & 73.90          & 75.80          & 64.38          \\
          DAEL~\cite{zhou2021domain}   & 59.40          & 55.10          & 74.00          & 75.70          & 66.10          \\
          FSDCL~\cite{jeon2021feature} & 60.24          & 53.54          & 74.36          & 76.66          & 66.20          \\
          FACT~\cite{xu2021fourier}    & \textbf{60.34} & \textbf{54.85} & 74.48          & 76.55          & 66.56          \\
          \midrule
          MVDG (Ours)                         & 60.25          & 54.32          & \textbf{75.11} & \textbf{77.52} & \textbf{66.80} \\
          \bottomrule
        \end{tabular}
        \label{tab:oh}
      }
    \end{minipage}
  \end{center}
  %\vspace{-20pt}
\end{table}

\textbf{PACS.} We perform evaluation on PACS with ResNet-18 and ResNet-50 as our backbone. We compare with several meta-learning based methods~(\ie, MLDG \cite{li2018learning}, MASF \cite{dou2019domain}, MetaReg \cite{balaji2018metareg}), augmentation based methods (\ie, FACT \cite{xu2021fourier}, FSDCL \cite{jeon2021feature}), ensemble learning based methods (\ie, SWAD \cite{cha2021swad}), domain-invariant feature learning (\ie, VDN \cite{wang2021variational}) and causal reasoning (\ie, MatchDG~\cite{mahajan2021domain}). As shown in \cref{tab:pacs}, our method can surpass traditional meta-learning methods in a large margin by 4.86\% (86.56\% vs. 81.70\%) on ResNet-18 and 6.66\% (89.33\% vs. 82.67\%) on ResNet-50. Besides, our method also achieves SOTA performance compared to recent other methods. Note that the improvement of our method on the hardest ``sketch'' domain is significant compared to the DeepAll~(\ie, 85.08\% vs. 66.21\%), owing to its better regularization and robustness.

\textbf{VLCS.} To verify the trained model can also generalize to unseen domains with a small domain gap, we conduct an experiment on VLCS. As seen in \cref{tab:vlcs}, our method outperforms several SOTA methods and achieves the best performance on three domains~(CALTECH, LABELME, and PASCAL), demonstrating that our method can also perform well in this case.

\textbf{OfficeHome.} We compare our method with SOTA methods on OfficeHome to prove the adaptation of our method to the dataset with a large number of classes. The result is reported in \cref{tab:oh}. Our method is able to achieve comparable performance to current SOTA methods.

\subsection{Ablation Study}

To further investigate our method, we conduct an ablation study on MVDG: \ie, 1) Reptile version of MLDG, 2) multi-view regularized meta-learning (MVRML), 3) multi-view prediction~(MVP).

\begin{table}[h]
  \caption{The accuracy~(\%) of the ablation study on PACS dataset on each component:  DeepAll, Reptile, and multi-view regularized  meta-learning~(MVRML), multi-view prediction~(MVP). }
  \begin{center}
    \renewcommand\arraystretch{1.2}
    \setlength\tabcolsep{2pt}
    \begin{minipage}{.49\columnwidth}
      \resizebox{1\columnwidth}{!}{
        \begin{tabular}{l | c c c c | c}
          \toprule
          ~\textbf{Method}~ & \textbf{A} & \textbf{C} & \textbf{P} & \textbf{S} & \textbf{Avg.}  \\
          \midrule
          Resnet-18         & 78.40      & 75.76      & 96.49      & 66.21      & 79.21$\pm$0.49 \\
         
          +MVRML            & 85.20      & 79.97      & 95.29      & 83.11      & 85.89$\pm$0.27 \\
          +MVRML+MVP        & 85.62      & 79.98      & 95.54      & 85.08      & 86.56$\pm$0.18 \\
          \midrule
          +Reptile          & 80.49      & 76.23      & 94.91      & 77.65      & 82.34$\pm$0.71 \\
          \bottomrule
        \end{tabular}
      }
    \end{minipage}
    \begin{minipage}{.49\columnwidth}
      \resizebox{1\columnwidth}{!}{
        \begin{tabular}{l | c c c c | c}
          \toprule
          ~\textbf{Method}~ & \textbf{A} & \textbf{C} & \textbf{P} & \textbf{S} & \textbf{Avg.}  \\
          \midrule
          Resnet-50         & 86.72      & 76.25      & 98.22      & 76.31      & 84.37$\pm$0.42 \\
          +MVRML            & 89.02      & 84.47      & 97.09      & 85.24      & 88.95$\pm$0.06 \\
          +MVRML+MVP        & 89.31      & 84.22      & 97.43      & 86.36      & 89.33$\pm$0.18 \\
          \midrule
          +Reptile          & 85.48      & 78.58      & 97.07      & 77.31      & 84.61$\pm$0.15 \\
          \bottomrule
        \end{tabular}
      }
    \end{minipage}
    \label{tab:ablation}

  \end{center}
  %\vspace{-20pt}
\end{table}

\textbf{Reptile.} As seen in \cref{tab:ablation}, the performance of Reptile can achieve satisfactory performance compared to DeepAll. Note that, although its performance in the ``sketch'' domain improves a lot~(\ie, 77.65\% vs. 66.21\%), the performance in the ``photo'' domain decreases. We hypothesize that the feature spaces learned by meta-learning and DeepAll are different. Since ResNet-18 is pretrained on ImageNet (photo-like dataset), it shows high performance in the ``photo'' domain at the beginning. When the training procedure continues, the model is hard to move far away from its initial weight space. Thus its performance is promising in the ``photo'' domain. However, when trained with meta-learning, it can obtain a good performance by the episodic training scheme but with a little sacrifice of its original performance in the ``photo'' domain.

\textbf{Multi-view regularized meta-learning.} When we apply multi-view regularized meta-learning~(MVRML), the performance is improved a lot on the baseline in \cref{tab:ablation}, which shows its efficacy in dealing with the overfitting issue. We observe that the ``photo'' domain also decreases a little. It may be caused by the better weight space produced by the meta-learning algorithm, which is far away from the initial weight space (\ie, the initial model trained by ImageNet). %It validates that MVRML can learn a better weight space.

\textbf{Multi-view prediction.} We employ multi-view prediction~(MVP) to enhance model reliability. As shown in \cref{tab:ablation}, the performance improves on both baselines. We notice that there is a large improvement in the  ``sketch'' domain because the  ``sketch'' domain has only the outline of the object, and thus it is more sensitive to small perturbations. With MVP, the model's prediction can be more reliable and accurate.

\subsection{Further Analysis}
\label{sec:exp_further}

We further analyze the properties of our method with ResNet-18 as backbone. More experiments could be found in \textbf{Supplementary Material}.

\begin{figure}[t]

  \resizebox{1\linewidth}{!}{
    \hspace{1pt}
    \begin{subfigure}{0.49\linewidth}
      \includegraphics[width=1\linewidth]{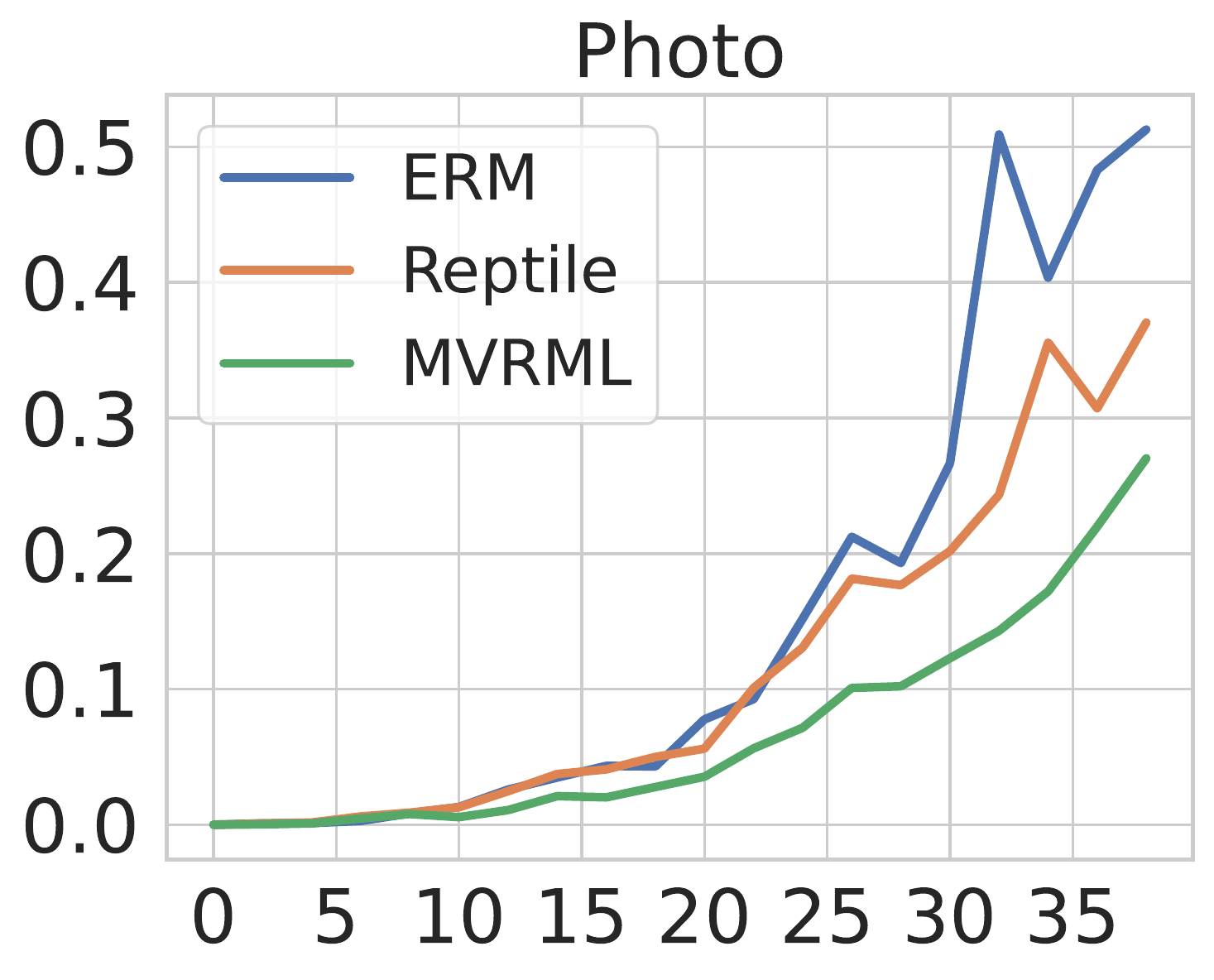}
    \end{subfigure}
    \hspace{1pt}
    \begin{subfigure}{0.49\linewidth}
      \includegraphics[width=1\linewidth]{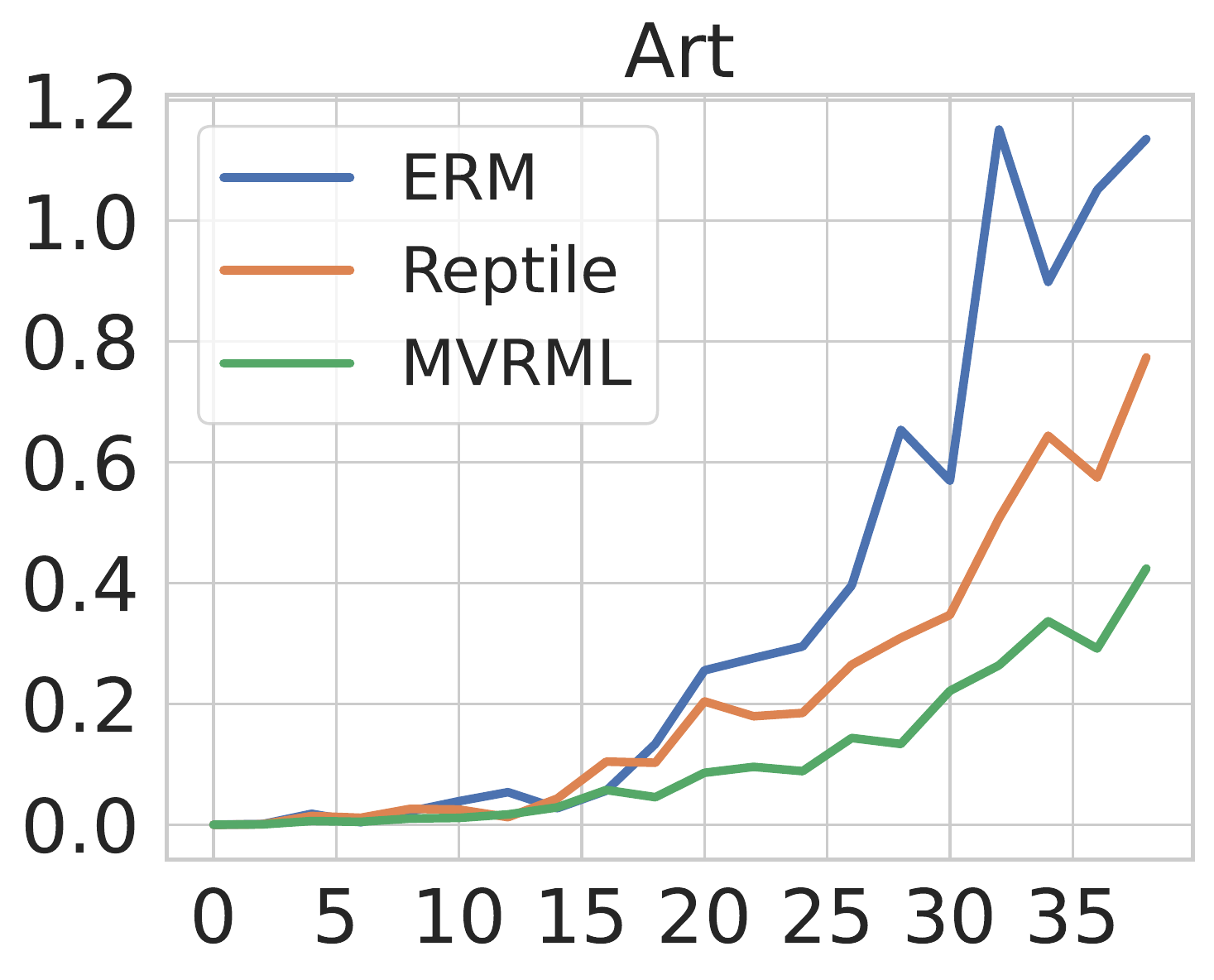}
    \end{subfigure}
    \hspace{1pt}
    \begin{subfigure}{0.49\linewidth}
      \includegraphics[width=1\linewidth]{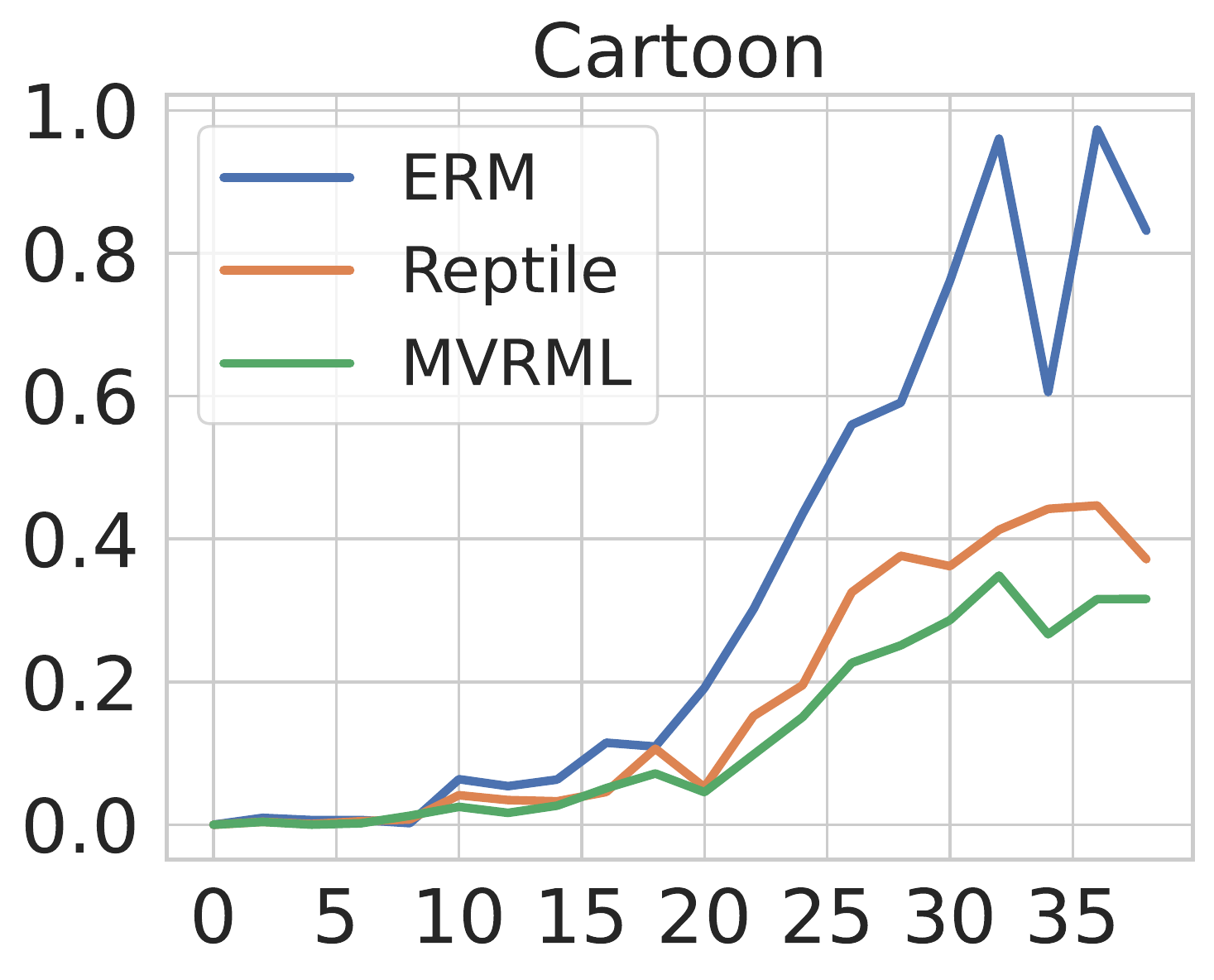}
    \end{subfigure}
    \hspace{1pt}
    \begin{subfigure}{0.49\linewidth}
      \includegraphics[width=1\linewidth]{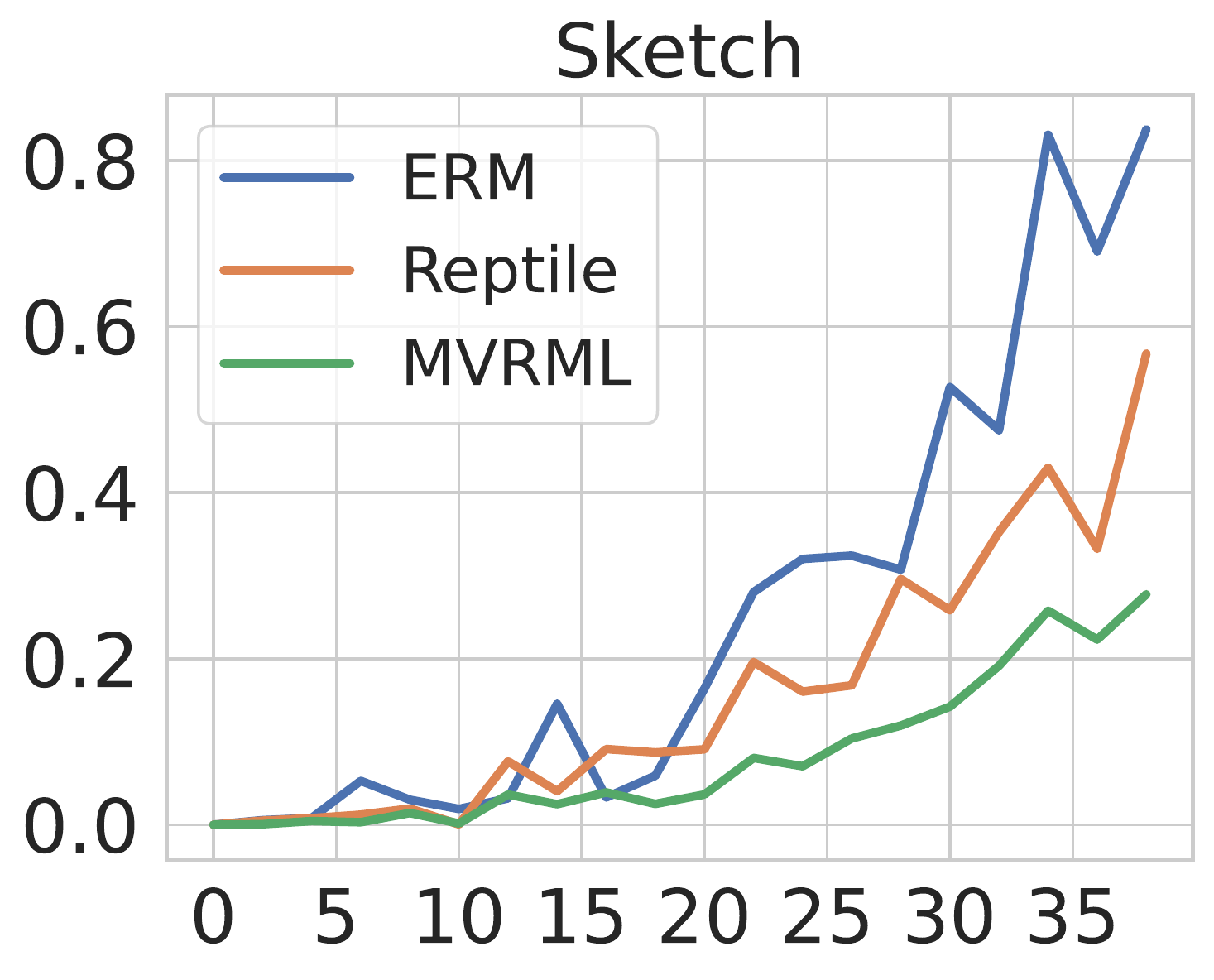}
    \end{subfigure}
  }

  \caption{Local sharpness comparison between ERM, Reptile, and MVRML on target domain of PACS. X-axis indicates the distance $\gamma$ to the original parameter, and Y-axis indicates the sharpness of the loss surface (the lower, the flatter).}
  %\vspace{-10pt}
  \label{fig:flat_test_appendix}
\end{figure}

\textbf{Local sharpness comparison.} As mentioned in \cref{Sec:MVRML}, our method can achieve better performance because it can find a flat minimum. To verify it, we plot local flatness via the loss gap between the original parameter and perturbed parameter, \ie, $\mathbb{E}_{\theta'=\theta+\epsilon}[\mathcal{L}(\theta, \mathcal{D}) - \mathcal{L}(\theta', \mathcal{D})]$~\cite{cha2021swad}, where $\epsilon$ is perturbation sampled from gaussian distribution with a radius of $\gamma$. We conduct an experiment on the target domain of PACS, and the results are shown in \cref{fig:flat_test_appendix}. With a larger radius, the sharpness of the three methods all increases. However, MVRML can find a better flat minimum where the loss increases the most slowly compared to the other methods, achieving better generalization performance.

\begin{table}[t]%{r}{0.40\linewidth}
  % \vspace{-25pt}
  \begin{center}
    \renewcommand\arraystretch{1.1}
    \caption{The comparison of different task sampling strategies for MVRML on PACS dataset.}
    %\vspace{-5pt}
    % \resizebox{0.4\columnwidth}{!}{
      \begin{tabular}{c | c c c c | c} 
        \toprule
        ~\textbf{Strategies}~ & \textbf{A} & \textbf{C} & \textbf{P} & \textbf{S} & \textbf{Avg.} \\
        \midrule
        S1                    & 82.81      & 76.85      & 94.65      & 82.78      & 84.27         \\
        S2                    & 84.11      & 78.48      & 93.75      & 82.95      & 84.82         \\
        S3~(Ours)             & 85.20      & 79.97      & 95.29      & 83.11      & 85.89         \\
        \bottomrule 
      \end{tabular}
      \label{tab:task_sampling}
    % }
  \end{center}
  % \vskip -20pt 
\end{table}
\textbf{Influence of task sampling strategies.}  {As mentioned in \cref{Sec:MVRML}, task sampling strategy is crucial for performance improvement in MVRML. We compare different sampling strategies at each iteration: each trajectory samples 1) from the same domain, denoted as S1; 2) from all domains, denoted as S2; 3) from random split meta-train and meta-test domains, denoted as S3.}
As shown in \cref{tab:task_sampling}, with a better sampling strategy, the generalizability of the trained model increases. We hypothesize that it is owing to 
the batch normalization layer that normalizes data with statistics calculated on a batch. When only sampling from a single domain, the diversity of BN statistics is limited to three domains. If we sample batches from different domains, although the diversity of BN statistics increases a little, the statistics in a batch tend to be the same on average. Finally, if we sample tasks from meta-train and meta-test splits, we can ensure that the diversity of statistics changes drastically, encouraging temporary models to explore diverse trajectories and produce a more robust optimization direction.

\begin{figure}[t]
  \centering

  \resizebox{1\linewidth}{!}{
    \begin{subfigure}{0.42\linewidth}
      \includegraphics[width=\linewidth]{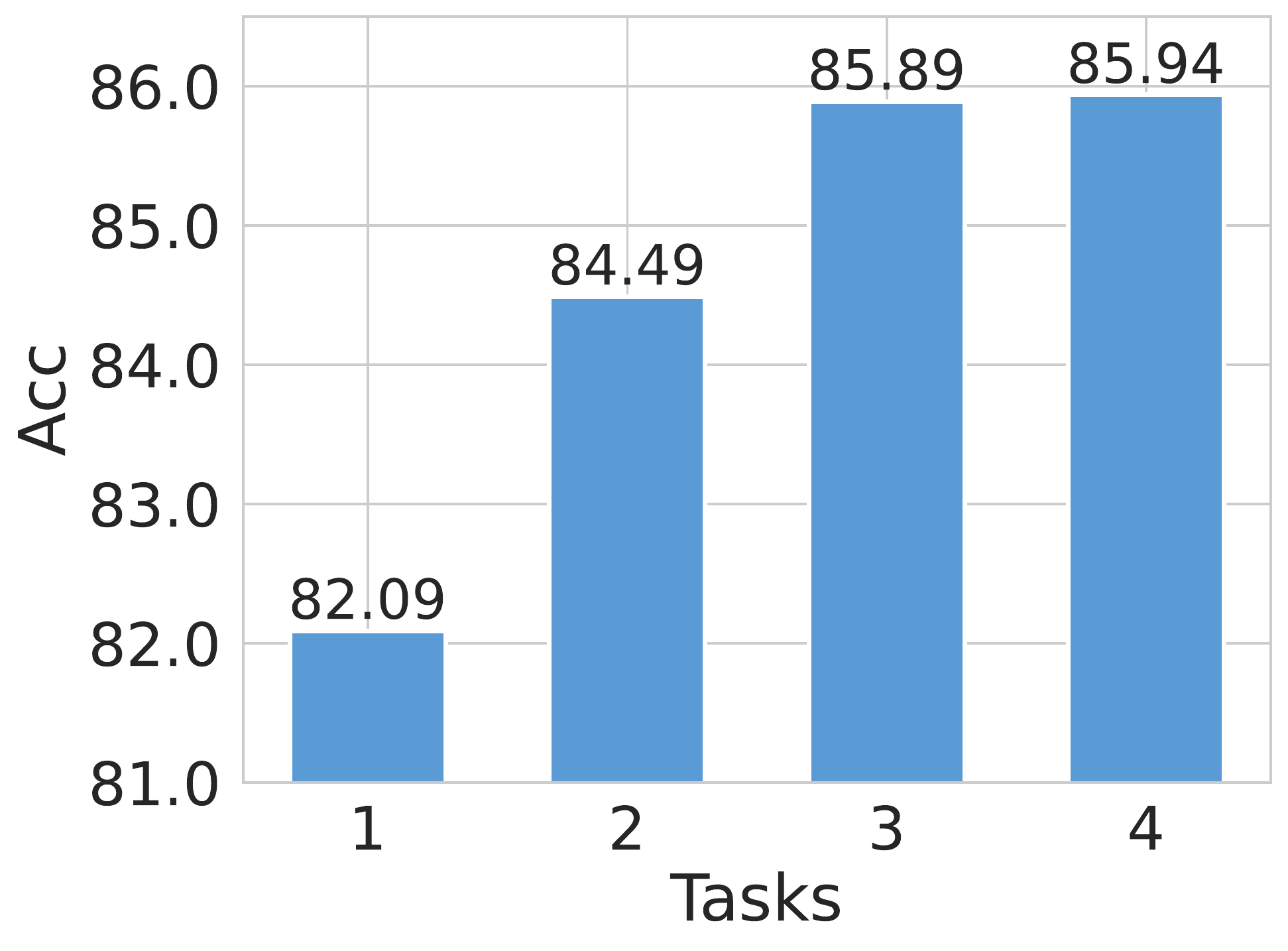}
      \caption{The number of tasks}
      \label{fig:task_number}
    \end{subfigure}
    \hspace{5pt}
    \begin{subfigure}{0.42\linewidth}
      \includegraphics[width=\linewidth]{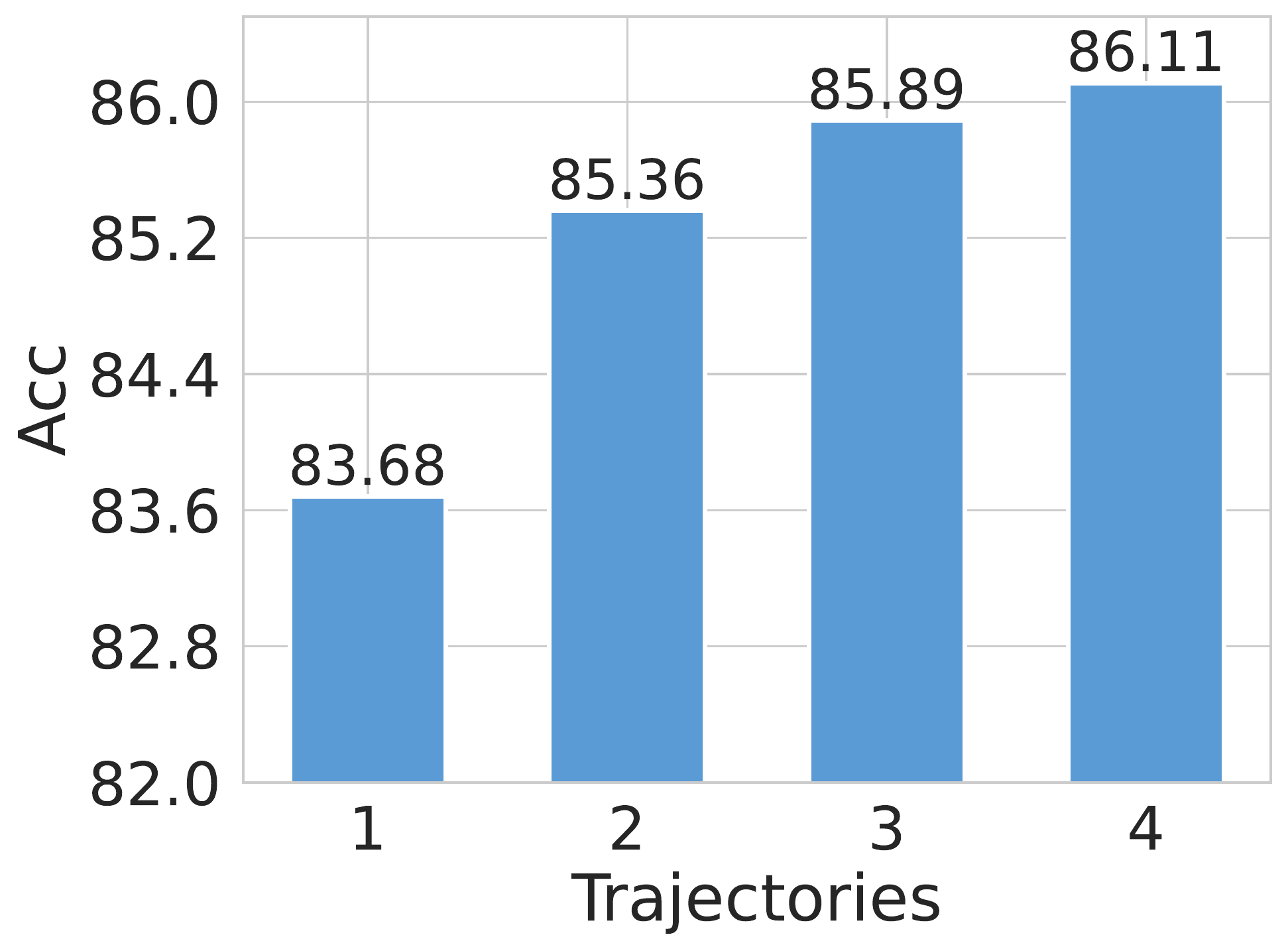}
      \caption{The number of trajectories}
      \label{fig:task_length}
    \end{subfigure}
    \hspace{5pt}
    \begin{subfigure}{0.5\linewidth}
      \includegraphics[width=1\linewidth]{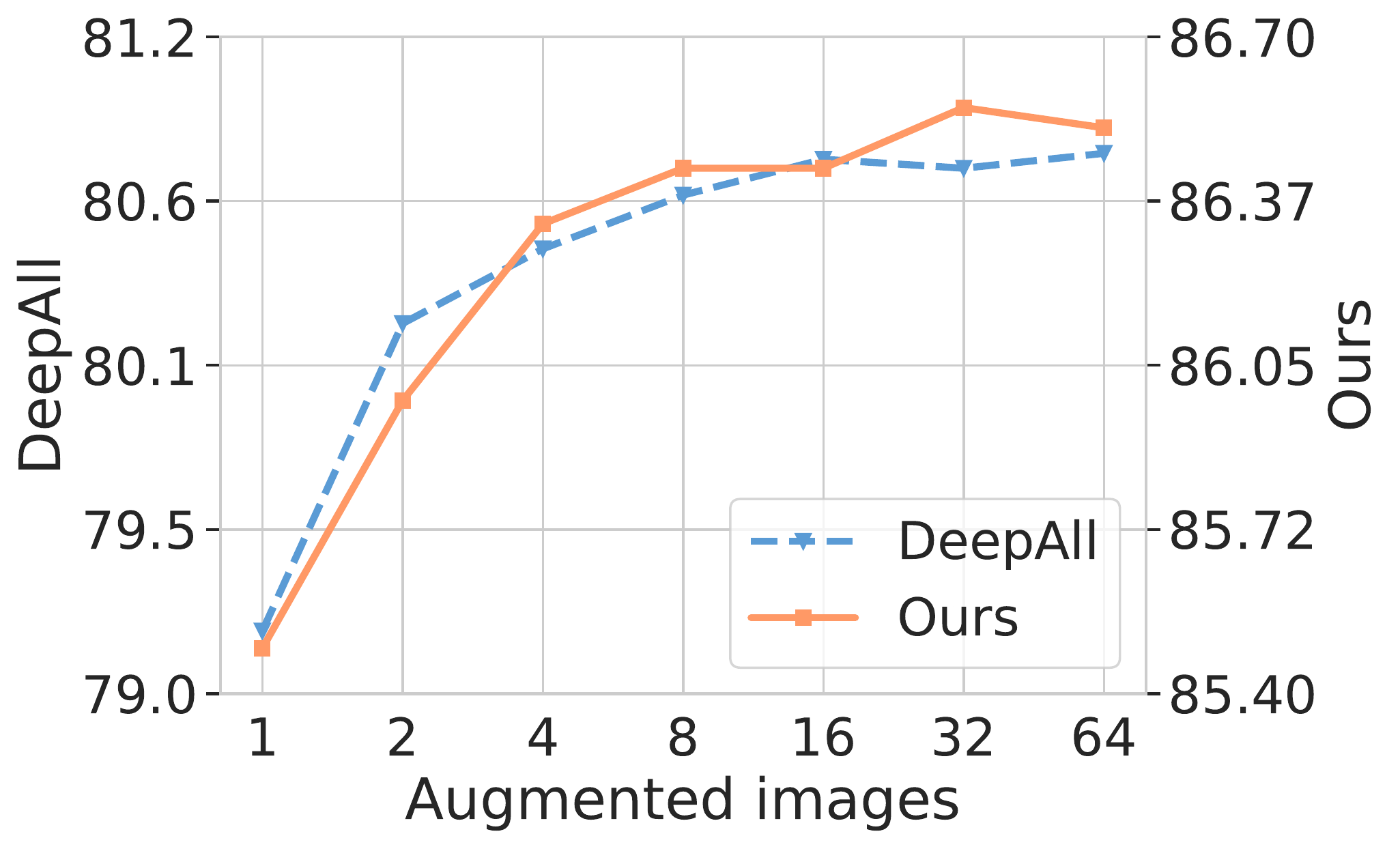}
      \caption{The number of augmented images}
      \label{fig:tta}
    \end{subfigure}
  }
  \caption{The impact of the number of tasks (a) and trajectories (b) in MVRML. The number of augmented images (c) in MVP.}
  %\vspace{-10pt}
\end{figure}

\textbf{Impact of the number of tasks and trajectories.} For MVRML, we sample a sequence of tasks on multiple trajectories to train the model. Both the number of tasks and trajectories can affect the generalization performance. Therefore, we train our model with a different number of tasks and trajectories. When we experiment with one factor, we set the other to $3$. The result can be shown in \cref{fig:task_number} and \cref{fig:task_length}. With the increasing number of tasks, the performance first improves, as our theory suggests that more tasks can benefit the generalizability.  However, the performance plateaus afterward. We suspect that the first few tasks are critical for the optimization procedure of temporary models since it decides the direction to optimize. With more tasks to be learned, the optimization direction does not change too much. Thus, more tasks could not largely improve performance. There is a similar trend in the number of trajectories. Because three trajectories are good enough to provide a robust optimization direction, more trajectories also could not help too much. To reduce the computational cost, we choose three trajectories in our experiments.

\textbf{Influence of the number of augmented images in MVP.} When applying multi-view prediction~(MVP), we need to augment the test images into different views and ensemble their results. Therefore, the number of augmented images has an influence on the result. We apply MVP to both DeepAll and our model with a different number of augmented images. As shown in \cref{fig:tta}, the performance improves when more augmented images are generated, which results from the increasing diversity of the test images. When the number is large enough (\eg, 64), the diversity created by the weak augmentation cannot increase anymore, and the performance plateaus.

\begin{table}[t]
  \footnotesize
  \caption{The left table shows accuracy and prediction change rate (PCR) of different methods on PACS dataset with ResNet-18. The weak augmentation is denoted as WA. The right table shows the accuracy~(\%) of applying MVP to other SOTA methods on PACS dataset. }
  \begin{center}
    \setlength\tabcolsep{2pt}
    \begin{minipage}{.52\columnwidth}
      \renewcommand\arraystretch{1.2}
      \resizebox{1\columnwidth}{!}{
        \begin{tabular}{c | l | cccc | c}
          \toprule
                                                            & \textbf{Method}  & \textbf{A} & \textbf{C} & \textbf{P} & \textbf{S} & \textbf{Avg.} \\
          \midrule
          \multirow{3}{*}{\rotatebox{90}{\textbf{Acc(\%)}}} & DeepAll~(w/o WA) & 77.90      & 70.60      & 96.41      & 66.03      & 77.73         \\
                                                            & DeepAll          & 78.40      & 75.76      & 96.49      & 66.21      & 79.21         \\
                                                            & MVRML       & 85.20      & 79.97      & 95.29      & 83.11      & 85.89         \\
          \midrule
          \multirow{3}{*}{\rotatebox{90}{\textbf{PCR(\%)}}} & DeepAll~(w/o WA) & 43.0       & 42.7       & 9.9        & 63.6       & 39.8          \\
                                                            & DeepAll          & 36.5       & 37.6       & 8.1        & 49.3       & 32.9          \\
                                                            & MVRML       & 27.7       & 28.4       & 8.7        & 25.4       & 22.6          \\
          \bottomrule
        \end{tabular}
        \label{tab:tta_pred}
      }
    \end{minipage}
    % \hspace{10pt}
    \begin{minipage}{.47\columnwidth}
      \renewcommand\arraystretch{1.2}
      \resizebox{1\columnwidth}{!}{
        \begin{tabular}{l | cccc |c}
          \toprule
          \textbf{Method}                  & \textbf{A} & \textbf{C} & \textbf{P} & \textbf{S} & \textbf{Avg.} \\
          \midrule
          MixStyle~\cite{zhou2021mixstyle} & 80.43      & 77.55      & 97.11      & 76.15      & 82.81         \\
          MixStyle+MVP                     & 83.05      & 78.45      & 97.50      & 78.99      & 84.49         \\
          \hline
          RSC~\cite{huang2020self}         & 84.67      & 76.75      & 95.81      & 82.59      & 84.96         \\
          RSC+MVP                          & 84.91      & 76.83      & 96.41      & 83.81      & 85.49         \\
          \hline
          FSR~\cite{wang2021feature}       & 84.57      & 80.76      & 96.11      & 82.16      & 85.90         \\
          FSR+MVP                          & 85.60      & 81.40      & 96.59      & 83.48      & 86.76         \\
          \bottomrule
        \end{tabular}
        \label{tab:tta_sota}
      }
    \end{minipage}
  \end{center}
  % %\vspace{-0.6cm}
  %\vspace{-20pt}
\end{table}

\textbf{Unstable prediction.} In the previous sections, we argue that if the model overfits the source domains, it is easy to produce unstable predictions by perturbing the test images slightly~(random resized crop and flip). By contrast, a robust model can reduce this effect and perform well. To verify it, we test several models in the unseen domain: DeepAll without weak augmentations (\ie, color jittering, random crop, and flip), DeepAll, and the model trained with MVRML. We introduce prediction change rate (PCR), calculated by the ratio of the number of predictions changed after applying the augmentations and the number of total predictions. We compare the test accuracy and PCR in \cref{tab:tta_pred}.  The larger this measure, the more unstable the model in the unseen domains. As seen, DeepAll without augmentation produces the highest PCR and lowest Acc because this model overfits source domains. Meanwhile, with data augmentation and a better training strategy, the performance of the model largely improves, and PCR decreases drastically.

\textbf{MVP on SOTA methods.} MVP is a plug-and-play method that can be easily adapted to other methods. To validate its adaptation ability, we integrate it into three SOTA methods, \ie, Mixstyle~\cite{zhou2021mixstyle}, RSC~\cite{huang2020self} and FSR~\cite{wang2021feature}. For RSC and FSR, we directly use their pre-trained model. For MixStyle, we implement it by ourselves. The result is shown in \cref{tab:tta_sota}. MVP can improve all of these trained models, which suggests its effectiveness in DG.

\section{Conclusion}

In this paper, to resist overfitting, with the observation that the performance in DG models can be benefited by task-based augmentation in training and sample-based augmentation in testing, we propose a novel multi-view framework to boost generalization ability and reduce unstable prediction caused by overfitting. Specifically, during training, we designed a multi-view regularized meta-learning algorithm. During testing, we introduced multi-view prediction to generate different views of a single image for the ensemble to stabilize its prediction. We provide theoretical proof that increasing the number of tasks can boost generalizability, and we also empirically verified that our method can help find a flat minimum that generalizes better. By conducting extensive experiments on three DG benchmark datasets, we validated the effectiveness of our method.
\\
\\
\noindent{\textbf{Acknowledgement.} This work was supported by NSFC Major Program (62192 783), CAAI-Huawei MindSpore Project (CAAIXSJLJJ-2021-042A), China Postdoctoral Science Foundation Project (2021M690609), Jiangsu Natural Science Foundation Project (BK20210224), and CCF-Lenovo Bule Ocean Research Fund.}

%%%%%%%%% REFERENCES
\bibliographystyle{splncs04}
\bibliography{egbib}

\begin{thebibliography}{10}
\providecommand{\url}[1]{\texttt{#1}}
\providecommand{\urlprefix}{URL }
\providecommand{\doi}[1]{https://doi.org/#1}

\bibitem{al2021data}
Al-Shedivat, M., Li, L., Xing, E., Talwalkar, A.: On data efficiency of
  meta-learning. In: AISTATS (2021)

\bibitem{ayhan2018test}
Ayhan, M.S., Berens, P.: Test-time data augmentation for estimation of
  heteroscedastic aleatoric uncertainty in deep neural networks. In: MIDL
  (2018)

\bibitem{balaji2018metareg}
Balaji, Y., Sankaranarayanan, S., Chellappa, R.: Metareg: Towards domain
  generalization using meta-regularization. In: NeurIPS (2018)

\bibitem{ben2007analysis}
Ben-David, S., Blitzer, J., Crammer, K., Pereira, F., et~al.: Analysis of
  representations for domain adaptation. In: NeurIPS (2007)

\bibitem{bousmalis2016domain}
Bousmalis, K., Trigeorgis, G., Silberman, N., Krishnan, D., Erhan, D.: Domain
  separation networks. In: NeurIPS (2016)

\bibitem{bousquet2002stability}
Bousquet, O., Elisseeff, A.: Stability and generalization. JMLR  (2002)

\bibitem{carlucci2019domain}
Carlucci, F.M., D'Innocente, A., Bucci, S., Caputo, B., Tommasi, T.: Domain
  generalization by solving jigsaw puzzles. In: CVPR (2019)

\bibitem{cha2021swad}
Cha, J., Chun, S., Lee, K., Cho, H.C., Park, S., Lee, Y., Park, S.: Swad:
  Domain generalization by seeking flat minima. arXiv  (2021)

\bibitem{chattopadhyay2020learning}
Chattopadhyay, P., Balaji, Y., Hoffman, J.: Learning to balance specificity and
  invariance for in and out of domain generalization. In: ECCV (2020)

\bibitem{deng2009imagenet}
Deng, J., Dong, W., Socher, R., Li, L.J., Li, K., Fei-Fei, L.: Imagenet: A
  large-scale hierarchical image database. In: CVPR (2009)

\bibitem{dou2019domain}
Dou, Q., de~Castro, D.C., Kamnitsas, K., Glocker, B.: Domain generalization via
  model-agnostic learning of semantic features. In: NeurIPS (2019)

\bibitem{finn2017model}
Finn, C., Abbeel, P., Levine, S.: Model-agnostic meta-learning for fast
  adaptation of deep networks. In: ICML (2017)

\bibitem{frankle2020linear}
Frankle, J., Dziugaite, G.K., Roy, D., Carbin, M.: Linear mode connectivity and
  the lottery ticket hypothesis. In: ICML (2020)

\bibitem{garipov2018loss}
Garipov, T., Izmailov, P., Podoprikhin, D., Vetrov, D.P., Wilson, A.G.: Loss
  surfaces, mode connectivity, and fast ensembling of dnns. In: NeurIPS (2018)

\bibitem{goodfellow2014generative}
Goodfellow, I., Pouget-Abadie, J., Mirza, M., Xu, B., Warde-Farley, D., Ozair,
  S., Courville, A., Bengio, Y.: Generative adversarial nets. In: NeurIPS
  (2014)

\bibitem{he2016deep}
He, K., Zhang, X., Ren, S., Sun, J.: Deep residual learning for image
  recognition. In: CVPR (2016)

\bibitem{hendrycks2019augmix}
Hendrycks, D., Mu, N., Cubuk, E.D., Zoph, B., Gilmer, J., Lakshminarayanan, B.:
  Augmix: A simple data processing method to improve robustness and
  uncertainty. arXiv  (2019)

\bibitem{huang2017arbitrary}
Huang, X., Belongie, S.: Arbitrary style transfer in real-time with adaptive
  instance normalization. In: ICCV (2017)

\bibitem{huang2020self}
Huang, Z., Wang, H., Xing, E.P., Huang, D.: Self-challenging improves
  cross-domain generalization. In: ECCV (2020)

\bibitem{izmailov2018averaging}
Izmailov, P., Podoprikhin, D., Garipov, T., Vetrov, D., Wilson, A.G.: Averaging
  weights leads to wider optima and better generalization. arXiv  (2018)

\bibitem{jeon2021feature}
Jeon, S., Hong, K., Lee, P., Lee, J., Byun, H.: Feature stylization and
  domain-aware contrastive learning for domain generalization. In: ACMMM (2021)

\bibitem{kouw2019review}
Kouw, W.M., Loog, M.: A review of domain adaptation without target labels.
  TPAMI  (2019)

\bibitem{krizhevsky2012imagenet}
Krizhevsky, A., Sutskever, I., Hinton, G.E.: Imagenet classification with deep
  convolutional neural networks. In: NeurIPS (2012)

\bibitem{lee2021test}
Lee, H., Lee, H., Hong, H., Kim, J.: Test-time mixup augmentation for
  uncertainty estimation in skin lesion diagnosis. In: MIDL (2021)

\bibitem{li2017deeper}
Li, D., Yang, Y., Song, Y.Z., Hospedales, T.M.: Deeper, broader and artier
  domain generalization. In: ICCV (2017)

\bibitem{li2018learning}
Li, D., Yang, Y., Song, Y.Z., Hospedales, T.M.: Learning to generalize:
  Meta-learning for domain generalization. In: AAAI (2018)

\bibitem{li2018domain}
Li, H., Jialin~Pan, S., Wang, S., Kot, A.C.: Domain generalization with
  adversarial feature learning. In: CVPR (2018)

\bibitem{li2021progressive}
Li, L., Gao, K., Cao, J., Huang, Z., Weng, Y., Mi, X., Yu, Z., Li, X., Xia, B.:
  Progressive domain expansion network for single domain generalization. In:
  CVPR (2021)

\bibitem{li2021simple}
Li, P., Li, D., Li, W., Gong, S., Fu, Y., Hospedales, T.M.: A simple feature
  augmentation for domain generalization. In: ICCV (2021)

\bibitem{li2018deep}
Li, Y., Tian, X., Gong, M., Liu, Y., Liu, T., Zhang, K., Tao, D.: Deep domain
  generalization via conditional invariant adversarial networks. In: ECCV
  (2018)

\bibitem{li2019feature}
Li, Y., Yang, Y., Zhou, W., Hospedales, T.M.: Feature-critic networks for
  heterogeneous domain generalizationx. arXiv  (2019)

\bibitem{liu2020shape}
Liu, Q., Dou, Q., Heng, P.A.: Shape-aware meta-learning for generalizing
  prostate mri segmentation to unseen domains. In: MICCAI (2020)

\bibitem{mahajan2021domain}
Mahajan, D., Tople, S., Sharma, A.: Domain generalization using causal
  matching. In: ICML (2021)

\bibitem{matsuura2020domain}
Matsuura, T., Harada, T.: Domain generalization using a mixture of multiple
  latent domains. In: AAAI (2020)

\bibitem{melas2021pixmatch}
Melas-Kyriazi, L., Manrai, A.K.: Pixmatch: Unsupervised domain adaptation via
  pixelwise consistency training. In: CVPR (2021)

\bibitem{molchanov2020greedy}
Molchanov, D., Lyzhov, A., Molchanova, Y., Ashukha, A., Vetrov, D.: Greedy
  policy search: A simple baseline for learnable test-time augmentation. arXiv
  (2020)

\bibitem{muandet2013domain}
Muandet, K., Balduzzi, D., Sch{\"o}lkopf, B.: Domain generalization via
  invariant feature representation. In: ICML (2013)

\bibitem{na2021fixbi}
Na, J., Jung, H., Chang, H.J., Hwang, W.: Fixbi: Bridging domain spaces for
  unsupervised domain adaptation. In: CVPR (2021)

\bibitem{nichol2018reptile}
Nichol, A., Schulman, J.: Reptile: a scalable metalearning algorithm. arXiv
  (2018)

\bibitem{nuriel2020permuted}
Nuriel, O., Benaim, S., Wolf, L.: Permuted adain: Enhancing the representation
  of local cues in image classifiers. arXiv  (2020)

\bibitem{piratla2020efficient}
Piratla, V., Netrapalli, P., Sarawagi, S.: Efficient domain generalization via
  common-specific low-rank decomposition. In: ICML (2020)

\bibitem{qiao2020learning}
Qiao, F., Zhao, L., Peng, X.: Learning to learn single domain generalization.
  In: CVPR (2020)

\bibitem{rahman2019correlation}
Rahman, M.M., Fookes, C., Baktashmotlagh, M., Sridharan, S.: Correlation-aware
  adversarial domain adaptation and generalization. PR  (2019)

\bibitem{seo2019learning}
Seo, S., Suh, Y., Kim, D., Han, J., Han, B.: Learning to optimize domain
  specific normalization for domain generalization. In: ECCV (2020)

\bibitem{shu2020prepare}
Shu, M., Wu, Z., Goldblum, M., Goldstein, T.: Prepare for the worst:
  Generalizing across domain shifts with adversarial batch normalization. arXiv
   (2020)

\bibitem{sypetkowski2020augmentation}
Sypetkowski, M., Jasiulewicz, J., Wojna, Z.: Augmentation inside the network.
  arXiv  (2020)

\bibitem{thrunlorien}
Thrun, S.: Lorien pratt, editors. 1998. learning to learn

\bibitem{torralba2011unbiased}
Torralba, A., Efros, A.A.: Unbiased look at dataset bias. In: CVPR (2011)

\bibitem{venkateswara2017deep}
Venkateswara, H., Eusebio, J., Chakraborty, S., Panchanathan, S.: Deep hashing
  network for unsupervised domain adaptation. In: CVPR (2017)

\bibitem{wang2020learning}
Wang, S., Yu, L., Li, C., Fu, C.W., Heng, P.A.: Learning from extrinsic and
  intrinsic supervisions for domain generalization. In: ECCV (2020)

\bibitem{wang2021feature}
Wang, Y., Qi, L., Shi, Y., Gao, Y.: Feature-based style randomization for
  domain generalization. arXiv  (2021)

\bibitem{wang2021variational}
Wang, Y., Li, H., Chau, L.P., Kot, A.C.: Variational disentanglement for domain
  generalization. arXiv  (2021)

\bibitem{xu2021fourier}
Xu, Q., Zhang, R., Zhang, Y., Wang, Y., Tian, Q.: A fourier-based framework for
  domain generalization. In: CVPR (2021)

\bibitem{xu2020robust}
Xu, Z., Liu, D., Yang, J., Raffel, C., Niethammer, M.: Robust and generalizable
  visual representation learning via random convolutions. arXiv  (2020)

\bibitem{yue2019domain}
Yue, X., Zhang, Y., Zhao, S., Sangiovanni-Vincentelli, A., Keutzer, K., Gong,
  B.: Domain randomization and pyramid consistency: Simulation-to-real
  generalization without accessing target domain data. In: ICCV (2019)

\bibitem{yue2021transporting}
Yue, Z., Sun, Q., Hua, X.S., Zhang, H.: Transporting causal mechanisms for
  unsupervised domain adaptation. In: ICCV (2021)

\bibitem{zhang2022generalizable}
Zhang, J., Qi, L., Shi, Y., Gao, Y.: Generalizable model-agnostic semantic
  segmentation via target-specific normalization. PR  (2022)

\bibitem{zhao2018adversarial}
Zhao, H., Zhang, S., Wu, G., Moura, J.M., Costeira, J.P., Gordon, G.J.:
  Adversarial multiple source domain adaptation. In: NeurIPS (2018)

\bibitem{zhao2020domain}
Zhao, S., Gong, M., Liu, T., Fu, H., Tao, D.: Domain generalization via entropy
  regularization. In: NeurIPS (2020)

\bibitem{zhou2020deep}
Zhou, K., Yang, Y., Hospedales, T.M., Xiang, T.: Deep domain-ad image
  generation for domain generalisation. In: AAAI (2020)

\bibitem{zhou2021domain}
Zhou, K., Yang, Y., Qiao, Y., Xiang, T.: Domain adaptive ensemble learning. TIP
   (2021)

\bibitem{zhou2021mixstyle}
Zhou, K., Yang, Y., Qiao, Y., Xiang, T.: Mixstyle neural networks for domain
  generalization and adaptation. arXiv  (2021)

\end{thebibliography}

\appendix

\clearpage

\title{MVDG: A Unified Multi-view Framework for Domain Generalization\\
    ---Appendix---
} % Replace with your title
\author{Jian Zhang\inst{1,2} \and Lei Qi\inst{3}$^{,*}$ \and Yinghuan Shi\inst{1,2}$^{,*}$ \and Yang Gao\inst{1,2}}
\institute{State Key Laboratory for Novel Software Technology, Nanjing University, China. \and
    National Institute of Healthcare Data Science, Nanjing University, China.\and
    School of Computer Science and Engineering, Southeast University, China.
}
\maketitle

\section{Additional Experiments}

\begin{figure}[h]
    \centering
    \includegraphics[width=0.4\linewidth]{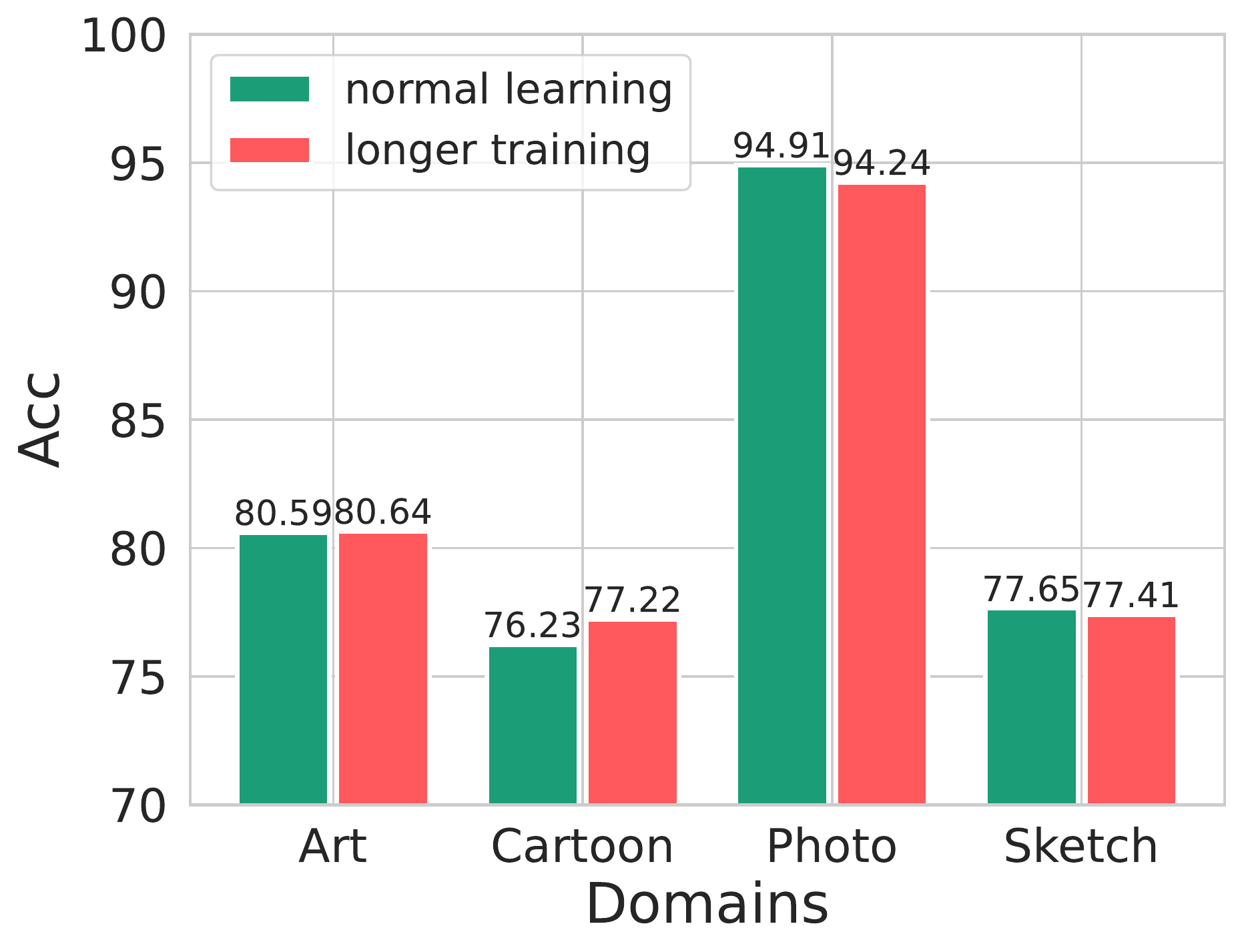}
    \caption{The comparison of accuracy (\%) of normal training and longer training in the meta-learning framework.}
    \label{fig:task_longer}
    %\vspace{-0.4cm}
\end{figure}

\textbf{Longer training.} In our training scheme, we need to train a model with more tasks than traditional meta-learning, resulting in a longer training time. To investigate whether it improves the performance, we train Reptile with a large epoch (\ie, 120 epochs). As shown in \cref{fig:task_longer}, the longer training does not bring any drastic performance gain compared to the original training epochs, which also validates the efficacy of our method.

\begin{figure}[h]
    \resizebox{1\linewidth}{!}{
        \hfill
        \begin{subfigure}{0.5\linewidth}
            \includegraphics[width=1\linewidth]{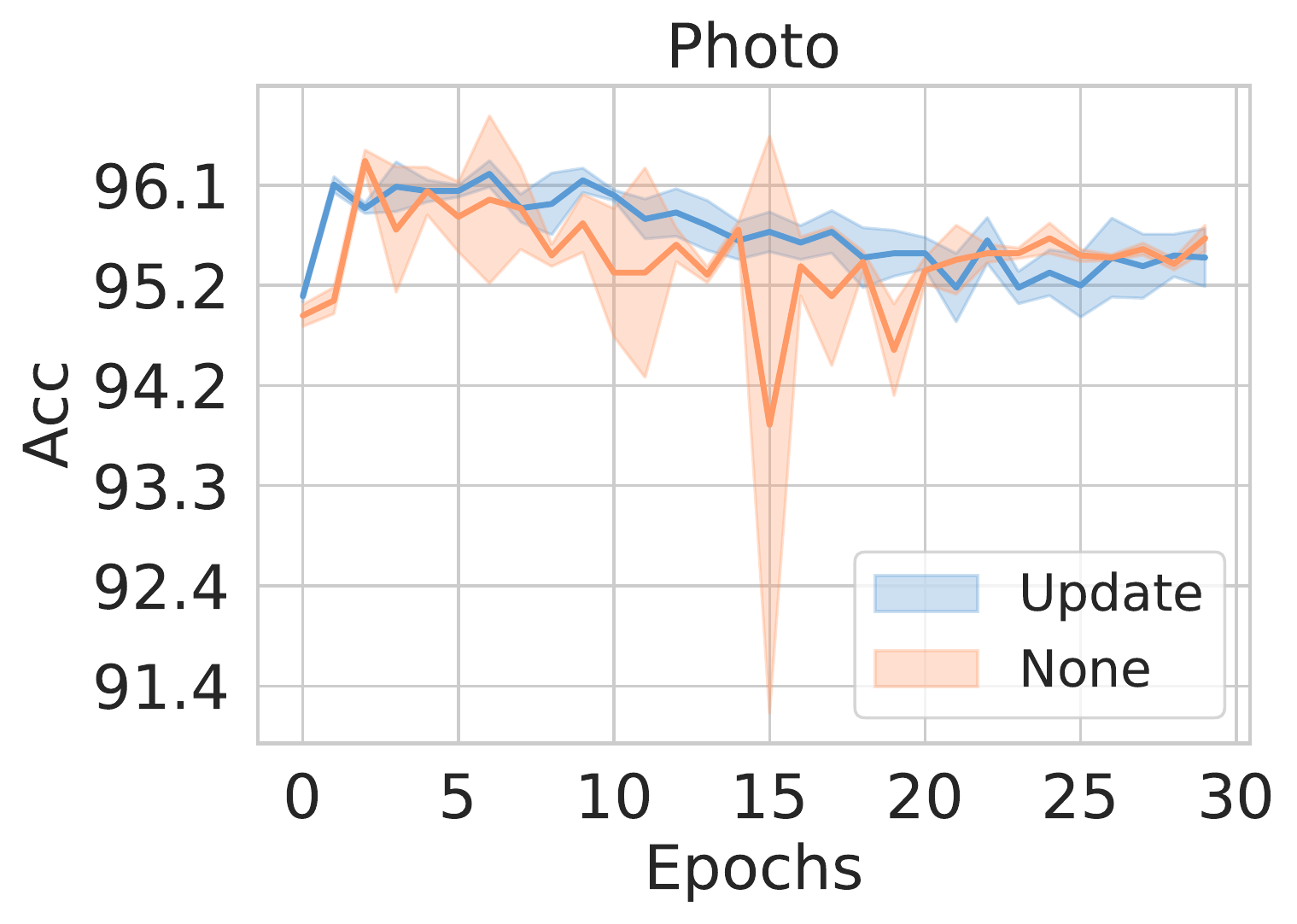}
        \end{subfigure}
        \hfill
        \begin{subfigure}{0.5\linewidth}
            \includegraphics[width=1\linewidth]{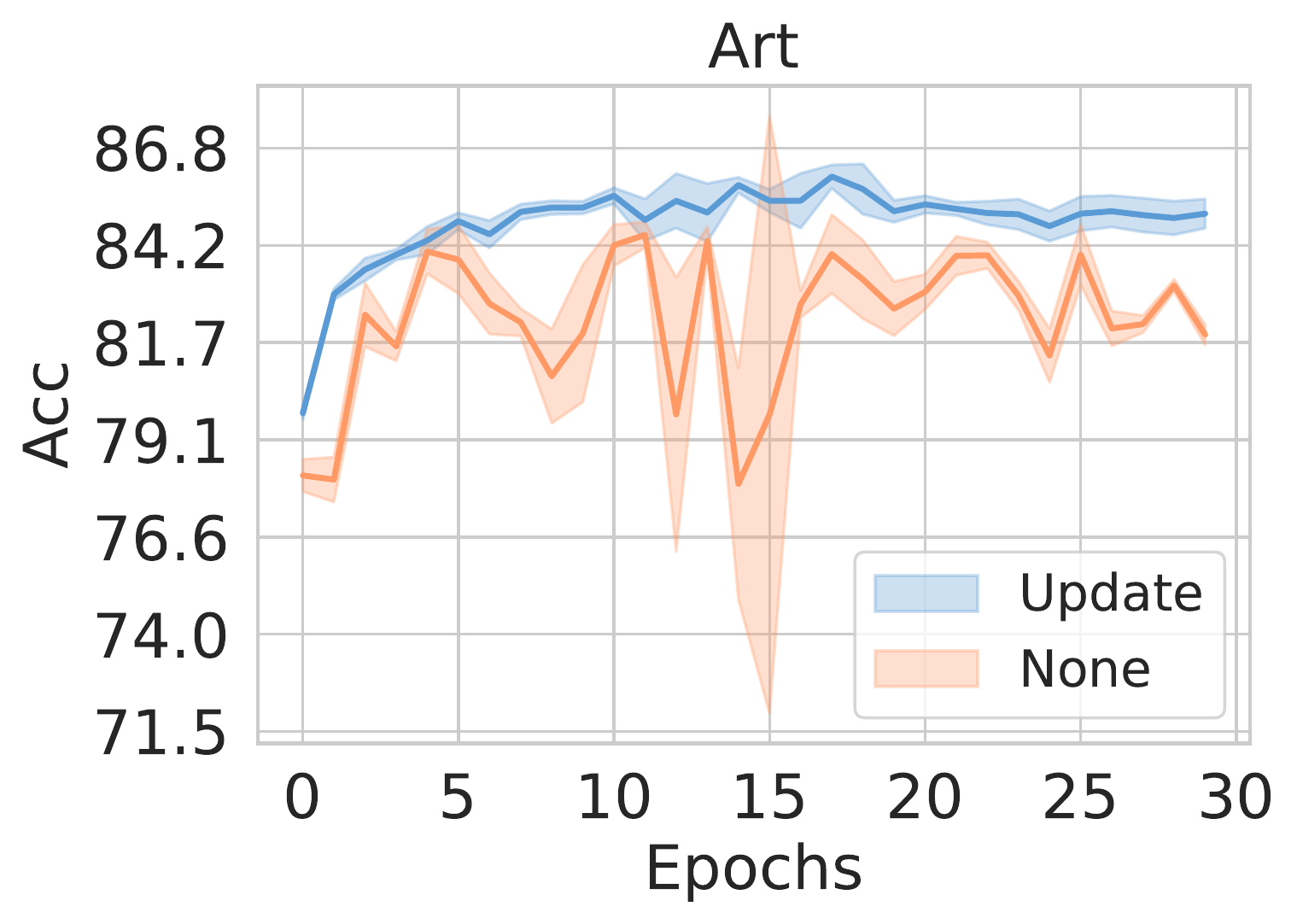}
        \end{subfigure}
        \hfill
        \begin{subfigure}{0.5\linewidth}
            \includegraphics[width=1\linewidth]{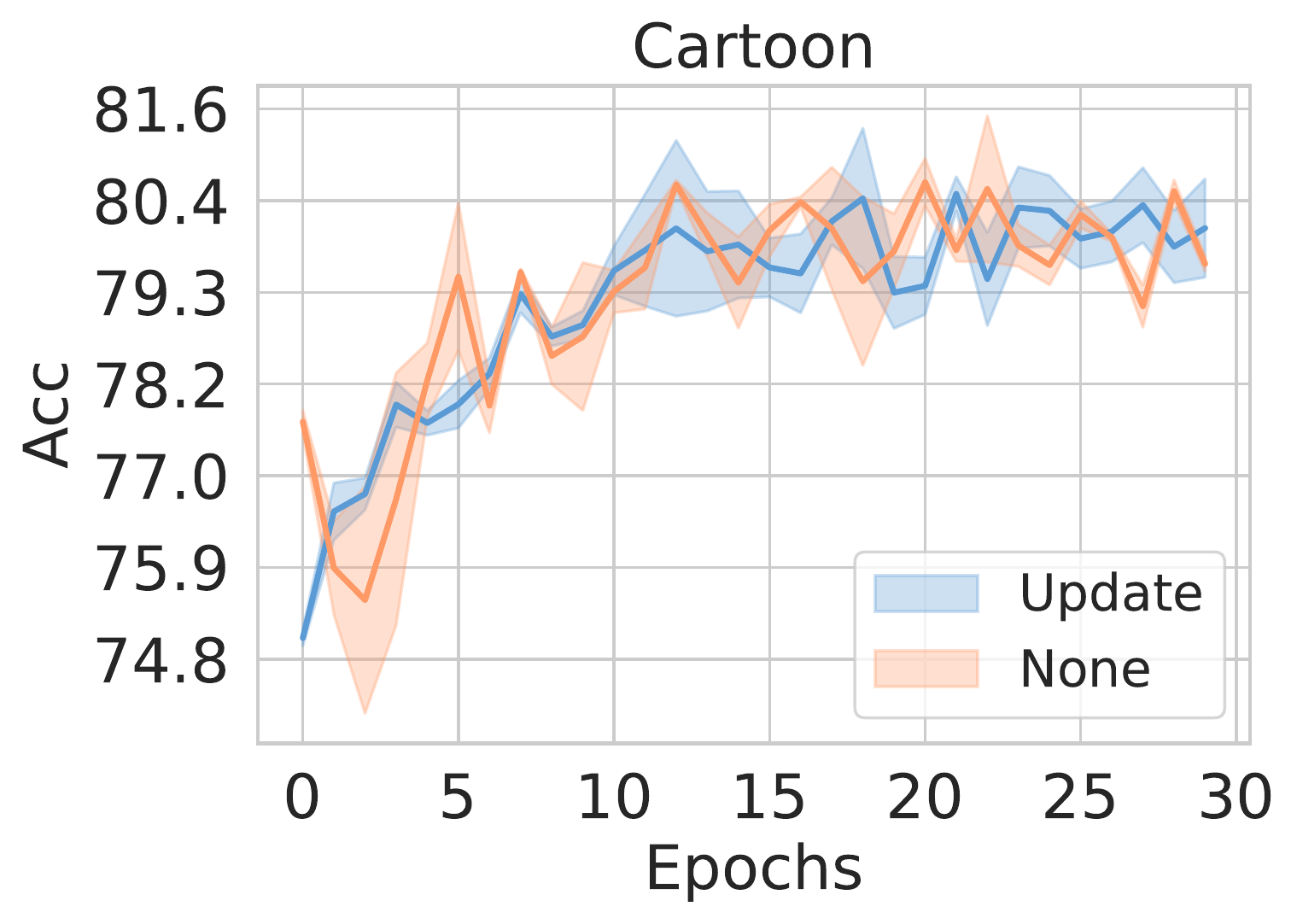}
        \end{subfigure}
        \hfill
        \begin{subfigure}{0.5\linewidth}
            \includegraphics[width=1\linewidth]{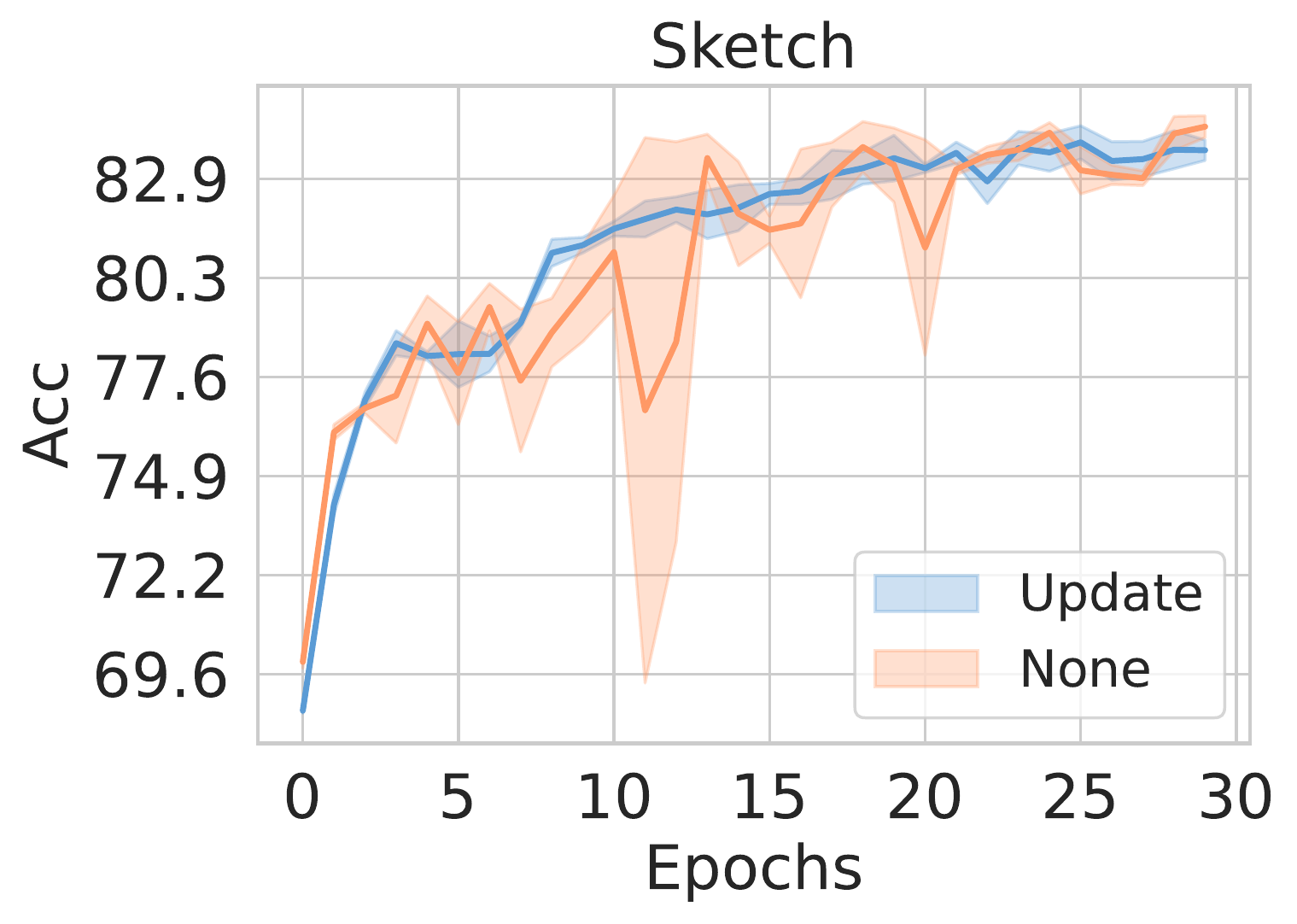}
        \end{subfigure}
        \hfill
    }
    \caption{The test accuracy (\%) of model with or without re-estimating BN statistics at eachepoch. The mean and standard deviation are denoted with the solid line and shaded areas, respectively.}
    %\vspace{-0.2cm}
    \label{fig:bn}
\end{figure}

\textbf{The effectiveness of re-estimating BN.} During the test stage, we note that the test accuracy is unstable, which is caused by the mismatch between BN statistics and model weights. During training, BN first normalizes data with statistics calculated in a batch and then keeps a running average of its statistics, which is used to normalize test images. However, for MVRML, the second step is not  {accomplished} because we only update model weights, leaving BN statistics unchanged, which causes a mismatch. Thus, the performance fluctuates when we apply these mismatched weights and statistics to test images.

A straightforward solution is to replace the statistics in the updated model with temporary model statistics, or we simply forward a current batch of data to continue accumulating the statistics in the updated model. However, we find that all these operations cannot help stabilize the training procedure. We hypothesize that the updating procedure $\theta_{j+1} = \theta_j + \beta (\theta_{tmp} - \theta_j)$ makes previous BN statistics totally unsuitable for the current weight and only a batch of data or replacement of statistics cannot remedy this effect. Therefore, at the end of the training stage, we need to re-estimate the statistics by forwarding the training set to find suitable statistics.

To visualize the performance fluctuation, we plot the mean~(the solid line) and standard deviation~(the shaded area) of model accuracy in the target domain. As seen on the orange line of Fig. \ref{fig:bn}, the performance of the model trained without re-estimating BN statistics fluctuates violently, making it hard to select the best model on the validation set. However, after we re-estimate BN statistics at the end of each epoch, its accuracy becomes more stable, and clear performance gains can be obtained in the ``art'' domain, as seen on the blue line.

% When we train the model using Reptile, the training process is unstable because the optimization procedure of Reptile is only performed on the model weights instead of the BN statistics, which causes a distribution mismatch between them. 
% We plot the mean~(the solid line) and standard deviation~(the shaded area) of model accuracy in the target domain. As seen on the orange line of Fig. \ref{fig:bn}, the performance of the model trained without re-estimating BN statistics fluctuates violently, making it hard to select the best model on the validation set. However, after we re-estimate BN statistics at the end of each epoch, its accuracy becomes more stable, and clear performance gains can be obtained in the ``art'' domain, as seen on the blue line.

\begin{figure}[h]

    \resizebox{1\linewidth}{!}{
        \hfill
        \begin{subfigure}{0.49\linewidth}
            \includegraphics[width=1\linewidth]{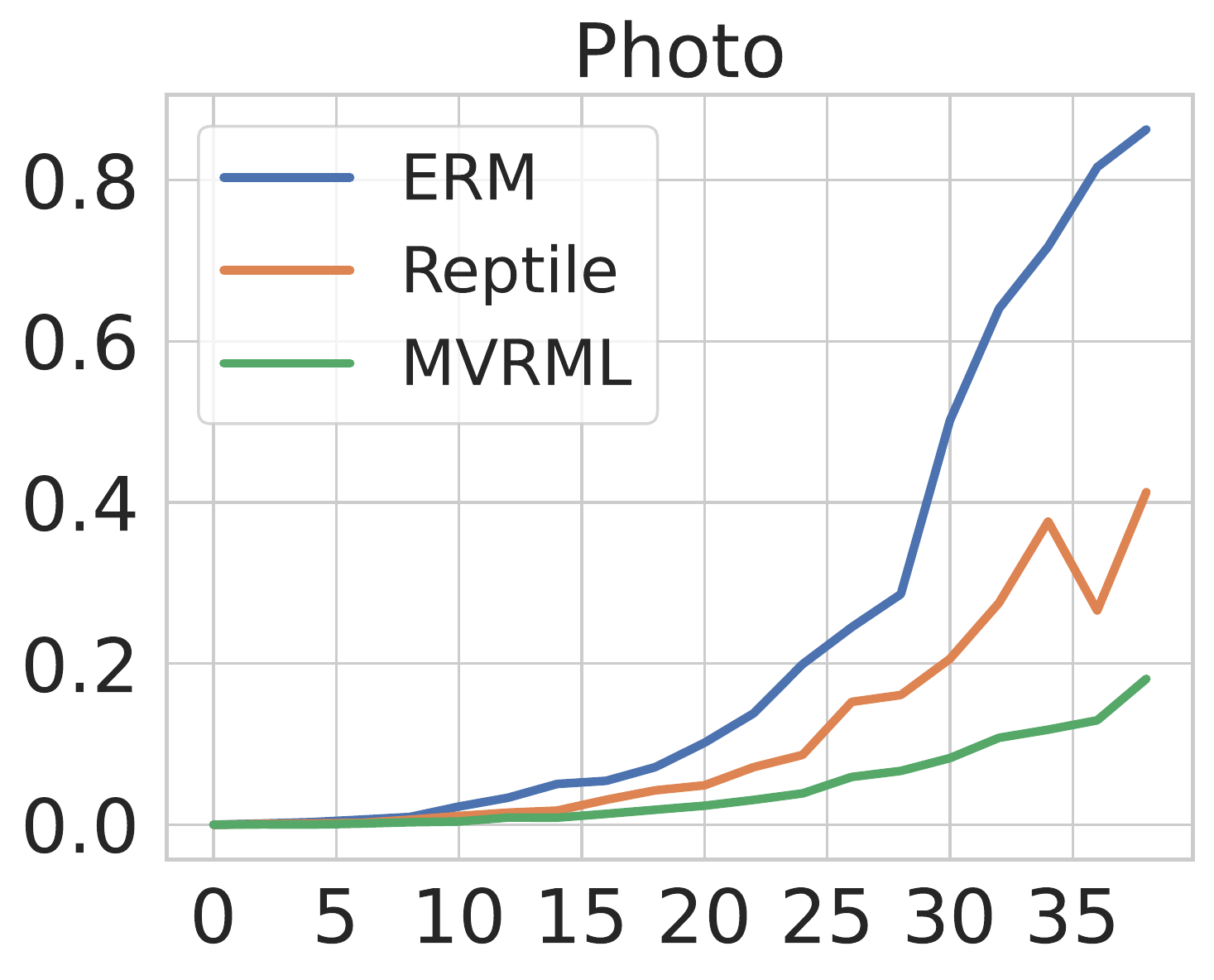}
        \end{subfigure}
        \hfill
        \begin{subfigure}{0.49\linewidth}
            \includegraphics[width=1\linewidth]{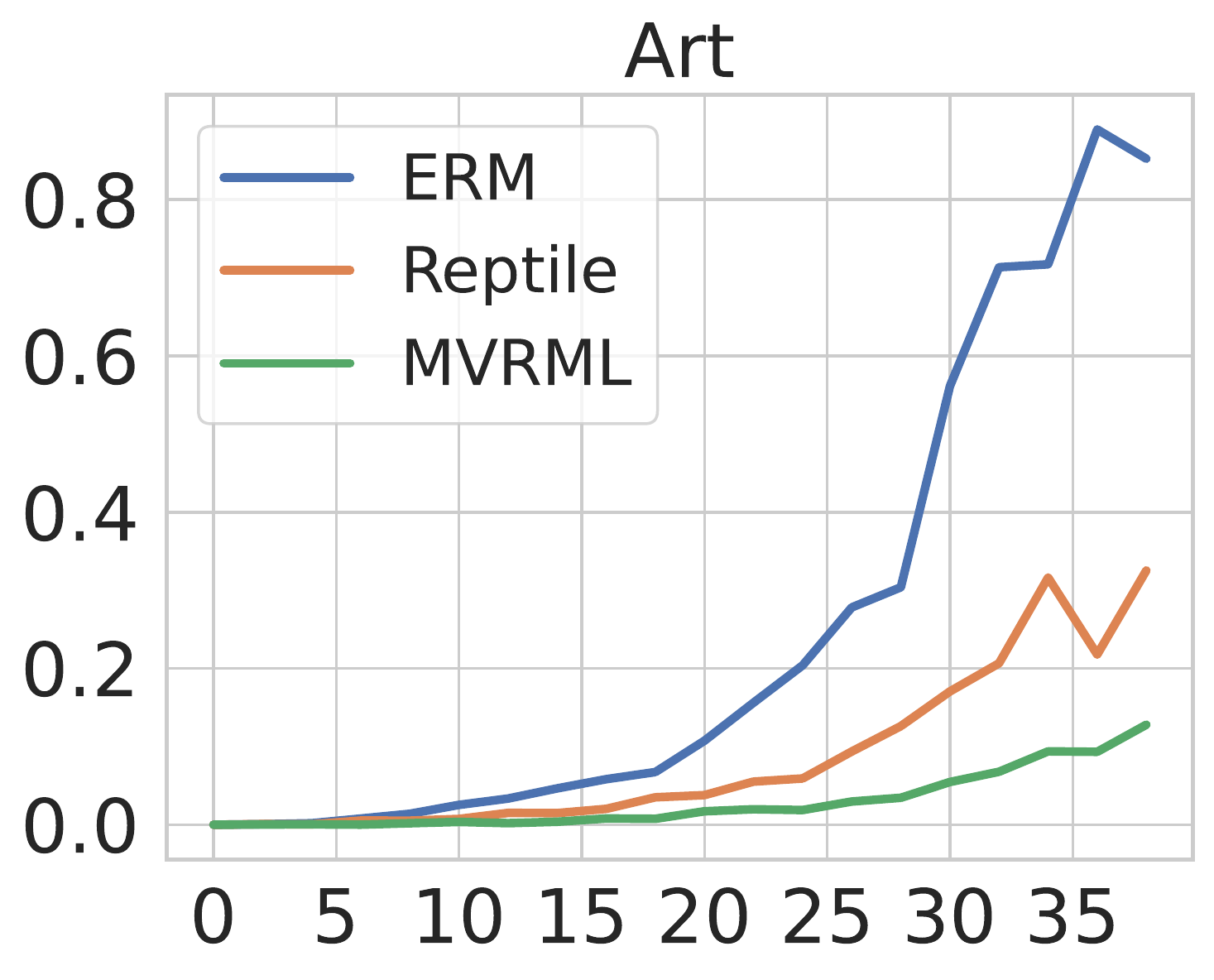}
        \end{subfigure}
        \hfill

        \hfill
        \begin{subfigure}{0.49\linewidth}
            \includegraphics[width=1\linewidth]{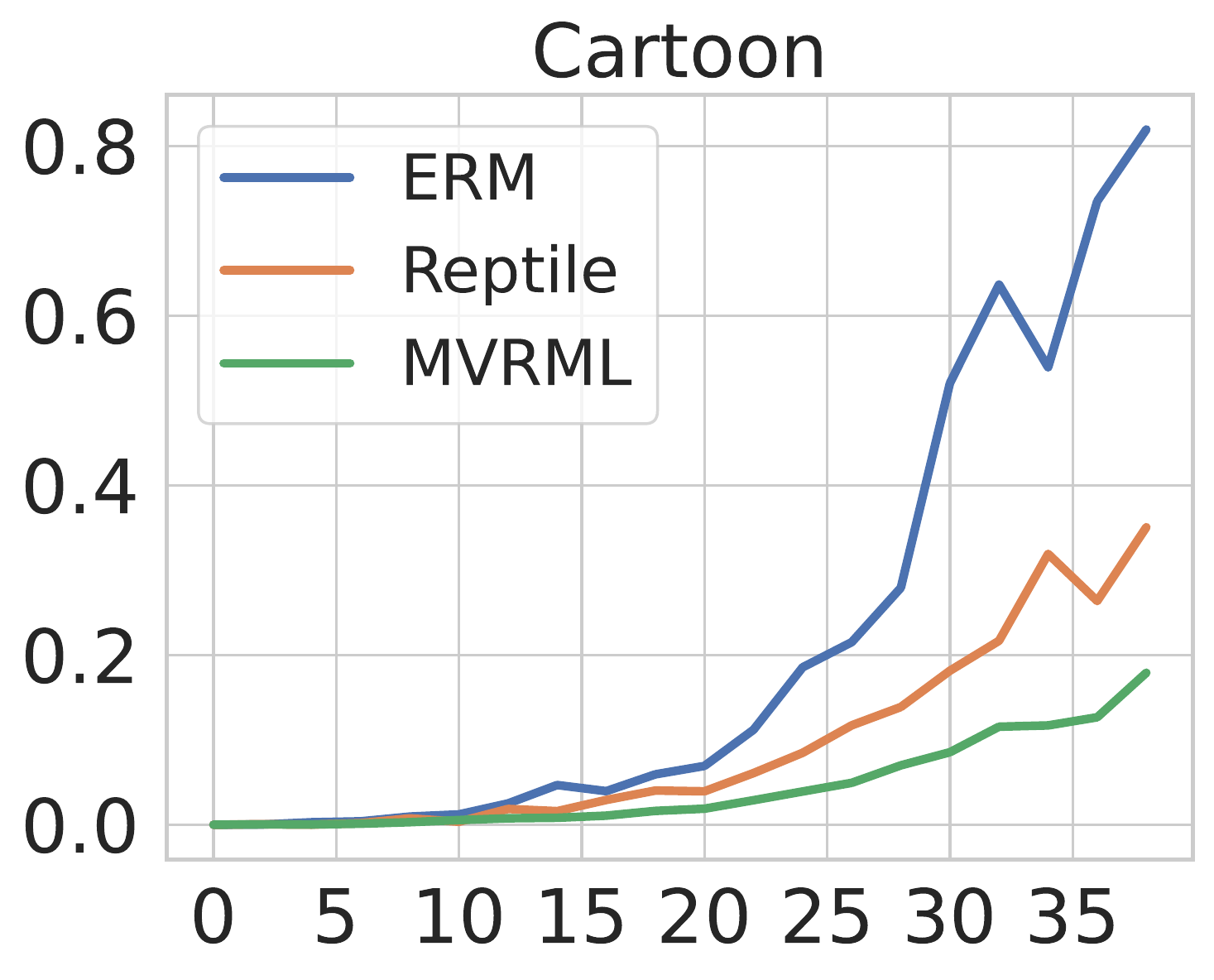}
        \end{subfigure}
        \hfill
        \begin{subfigure}{0.49\linewidth}
            \includegraphics[width=1\linewidth]{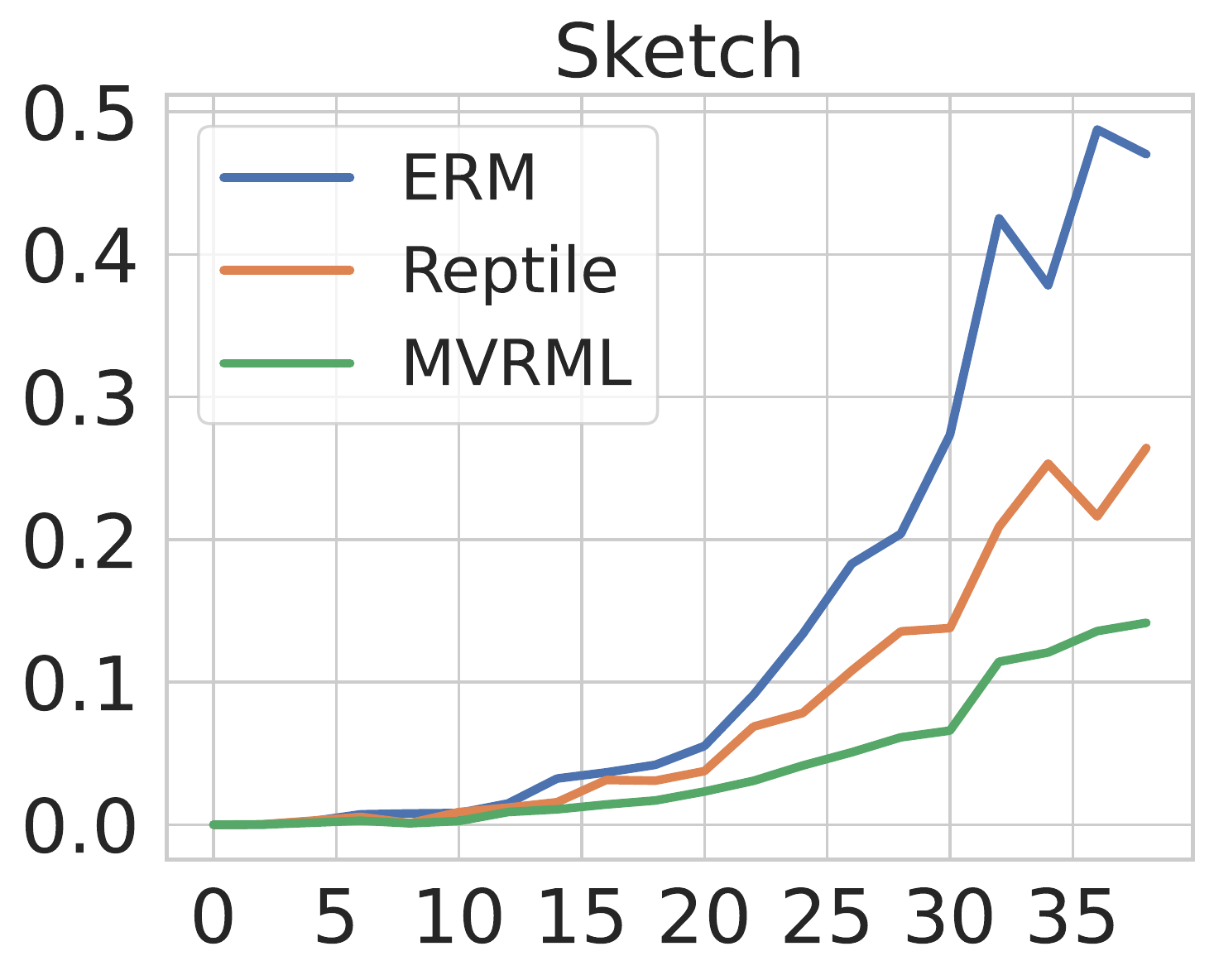}
        \end{subfigure}
        \hfill
    }

    \caption{Local sharpness comparison across ERM, Reptile and MVRML on the \textit{validation} set of PACS. The X-axis indicates the distance $\gamma$ to the original parameter and Y-axis indicates the sharpness of loss surface (the lower and stable, the more flat).}
    %\vspace{-0.2cm}
    \label{fig:flat_train_appendix}
\end{figure}

\textbf{Local sharpness comparison.} As mentioned in Sec. 4.4 in the main page that the sharpness is calculated by the gap between the original parameter and perturbed parameter, \ie, $\mathbb{E}_{\theta'=\theta+\epsilon}[\mathcal{L}(\theta, \mathcal{D}) - \mathcal{L}(\theta', \mathcal{D})]$. We sample the perturbation $10$ times from Gaussian distribution with different $\gamma$ and average the result to produce a stable sharpness value. In addition to the sharpness on the test set shown in Fig. 4 in the main page, we also plot the local sharpness on the validation set, as shown in \cref{fig:flat_train_appendix}. Our method also can find a flatter minimum than DeepAll and Reptile.

\begin{table}
    %\vspace{-25pt}
    \caption{The influence on accuracy~(\%) of different augmentation combination of multi-view prediction on PACS dataset. The best performance is marked as \textbf{bold}.}
    \begin{center}
        \renewcommand\arraystretch{1.1}
        \resizebox{0.5\columnwidth}{!}{
            \begin{tabular}{c c c c | cccc | c}
                \toprule
                \textbf{crop} & \textbf{flip} & \textbf{jitter} & \textbf{RA} & \textbf{A}     & \textbf{C}     & \textbf{P}     & \textbf{S}     & \textbf{Avg.}  \\
                \midrule
                              &               &                 &             & 85.20          & 79.97          & 95.29          & 83.11          & 85.89          \\
                \checkmark    &               &                 &             & 85.20          & 79.58          & 95.54          & 84.65          & 86.24          \\
                \checkmark    & \checkmark    &                 &             & \textbf{85.62} & 79.98          & \textbf{95.54} & \textbf{85.08} & \textbf{86.56} \\
                \checkmark    & \checkmark    & \checkmark      &             & 84.17          & \textbf{80.13} & 94.84          & 84.28          & 85.85          \\
                \checkmark    &               &                 & \checkmark  & 72.38          & 74.28          & 91.94          & 77.38          & 79.00          \\
                \checkmark    & \checkmark    &                 & \checkmark  & 72.35          & 74.09          & 91.80          & 77.63          & 78.97          \\
                \checkmark    & \checkmark    & \checkmark      & \checkmark  & 70.83          & 73.98          & 91.31          & 77.55          & 78.42          \\
                \bottomrule
            \end{tabular}
            \label{tab:tta_aug}
        }\\
    \end{center}
    %\vspace{-20pt}
\end{table}

\textbf{Weak vs. Strong augmentation in MVP.}
When we apply MVP, augmentation also plays a significant role in ensemble performance.
To investigate how different augmentation transformations affect performance, we select several weak and strong augmentations, including random resized crop with a scale factor of $[0.8, 1]$, random horizontal flip, color jittering with a magnitude of $0.4$, and RandAugment with $N=4$ and $M=5$. The model is trained with MVRML. The number of augmented images is set to 32. In \cref{tab:tta_aug}, simple weak augmentations~(\ie, random resized crop and flip) can achieve the best performance. By contrast, the strong augmentations~(\ie, the color jittering and RandAugment (RA)) have an adverse effect because the images augmented with strong augmentation drift off the data manifold~\cite{hendrycks2019augmix}, making it harder to predict.

\begin{figure}[h]
    %\vspace{-15pt}
    \begin{center}
        \includegraphics[width=0.98\linewidth]{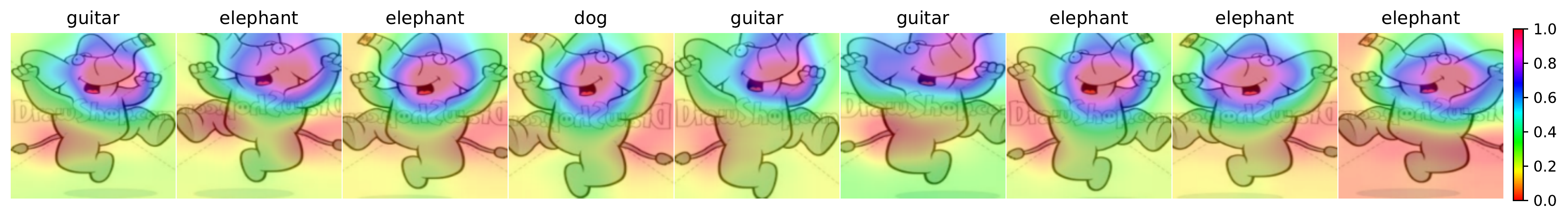}
    \end{center}
    %\vspace{-15pt}
    \caption{Class activation map of weak augmented images. The texts above the images are model predictions (best viewed in color).}
    %\vspace{-15pt}
    \label{fig:MVP_CAM}
\end{figure}

\textbf{Visualization of applying multi-view prediction.}  We visualize the class activation maps of weak augmented images. As shown in \cref{fig:MVP_CAM}, with small perturbations applied to the images, the model tends to make different predictions. However, by ensembling their predictions, the model could eliminate the situation that it makes a wrong prediction from a rare view of the testing image.

\textbf{The optimal number of trajectories in Reptile.} We compare Reptile by varying its number of trajectories. Specifically, the results of using 1, 2, 3, and 4 trajectories are 82.34\%, 82.84\%, 83.06\%, and 82.83\%, respectively. We notice that using 3 trajectories is optimal for Reptile. Since our method achieves 85.89\% \textit{without} multi-view prediction, its performance still can surpass Reptile.

\begin{table}[h]
    %\vspace{-20pt}
    \begin{center}
        \renewcommand\arraystretch{1}
        \caption{Comparison to SOTA with MVP on PACS dataset.}
        \begin{tabular}{r | l l l l l l}
            \toprule
                    & MLDG$^\dagger$ & FSDCL$^*$     & \textbf{MVDG} \\
            \midrule
            w/o MVP & 82.34          & 85.85         & 85.89         \\
            w/ MVP  & 83.41~(+1.07)  & 86.37~(+0.52) & 86.56~(+0.67) \\
            \bottomrule
        \end{tabular}
        \label{tab:mvp_on_sota}
    \end{center}
    %\vspace{-20pt}
\end{table}

\textbf{Comparison to SOTA with multi-view prediction.} We compare our method to other two SOTA methods (\ie, MLDG and FDSCL) equipped with MVP. We directly utilize MVP to their available trained models (denoted as *) or our reproduced models (denoted as $\dagger$).
As shown in \cref{tab:mvp_on_sota}, our method still outperforms these methods. Also, it validates the efficacy of MVP again.

\begin{table}[t]
    % %\vspace{-25pt}
    \begin{center}
        \renewcommand\arraystretch{1.0}
        \caption{The training time w.r.t. the number of tasks and trajectories on Photo domain.}

        \begin{tabular}{l | c c c cccccc }
            \toprule
            Tasks--Traj.  & 1-1 & 1-2 & 1-3 & 2-1 & 2-2 & 2-3 & 3-1 & 3-2 & 3-3 \\
            \midrule
            Time~(minute) & 11  & 14  & 16  & 16  & 22  & 27  & 21  & 26  & 32  \\
            \bottomrule
        \end{tabular}
        \label{tab:training_time}

    \end{center}
\end{table}

\begin{figure}[t]
    %\vspace{-10pt}
    \begin{center}
        \includegraphics[width=0.5\linewidth]{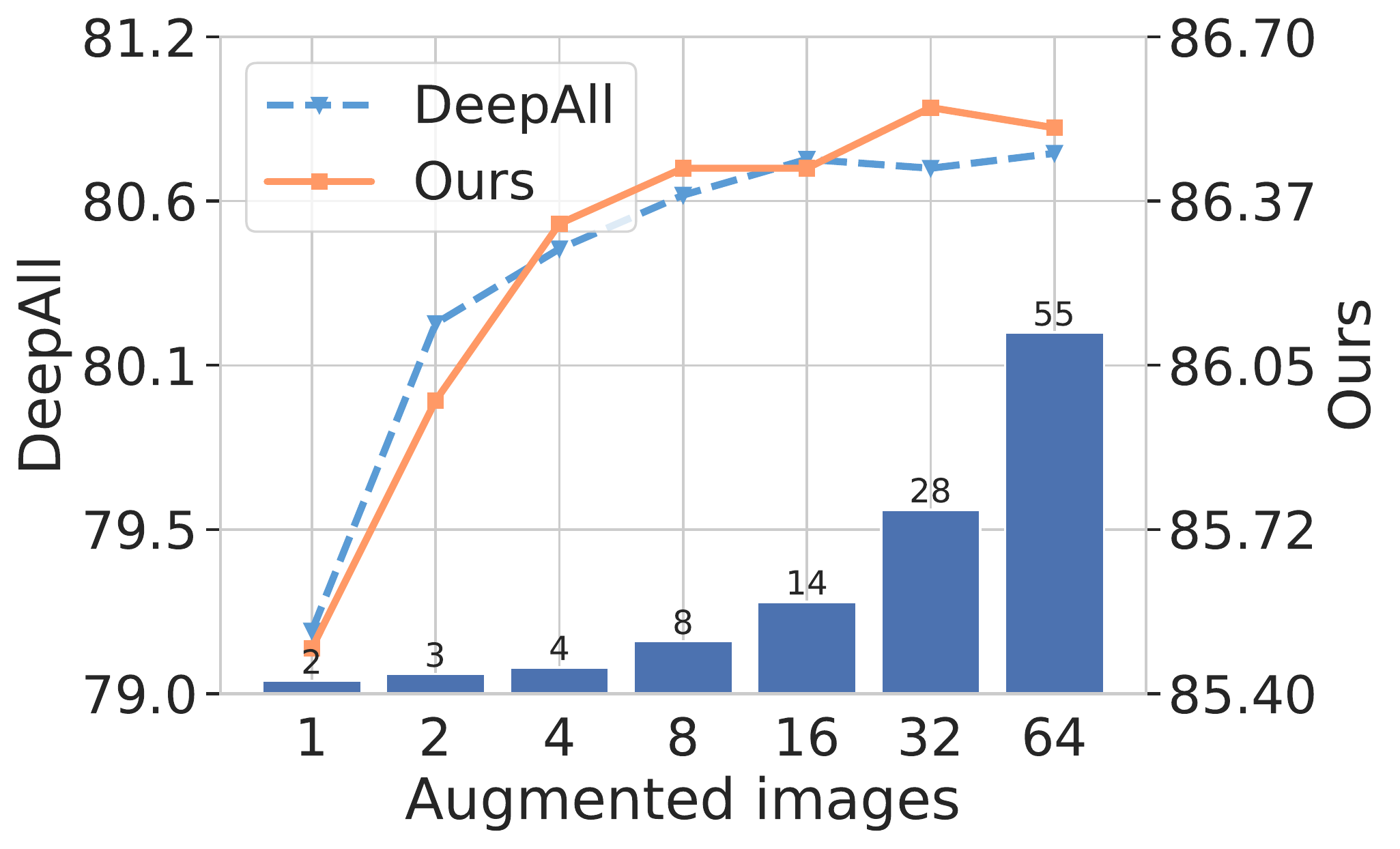}
    \end{center}
    %\vspace{-15pt}
    \caption{Accuracy(\%) (lines) and testing time(s) (histogram) with augmented images.}
    %\vspace{-20pt}
    \label{fig:testing_time}
\end{figure}

\textbf{Training and testing time of the method.} The experiments are all conducted on 2080Ti GPU. As shown in \cref{tab:training_time}, the training time of our method increases with more tasks and trajectories.
% The algorithm is implemented by parallelly training each trajectory.
Although our method requires 32 minutes to train a model, it is still comparable with SOTA (\eg, FACT: 2.81h, for the same epochs). % RSC: 13m,  
% Also, we claim that the training time does not influence the testing stage.
Also, we plot the testing time of MVP in \cref{fig:testing_time} on the Photo domain. The testing time increases with more augmented images. However, we find that 8 augmented images are enough to produce satisfying performance, which does not bring too much computational overhead. Besides, with 8 augmented images, the model can process 208 images per second on a 2080Ti card, which is sufficient for real-time tasks.

\begin{table}[h]
        \caption{Experiments on DomaniBed.}
    \begin{center}
        %\vspace{-20pt}
        \renewcommand\arraystretch{1}
        \begin{tabular}{l | c c c c c c c| c}
            \toprule
                          & CMNIST       & RMNIST       & VLCS         & PACS         & OfficeHome   & TerraInc     & DomainNet    & Avg   \\
            \midrule
            ERM           & 51.5$\pm$0.1 & 98.0$\pm$0.0 & 77.5$\pm$0.4 & 85.5$\pm$0.2 & 66.5$\pm$0.3 & 46.1$\pm$1.8 & 40.9$\pm$0.1 & 66.6  \\
            ERM$^*$       & 51.5$\pm$0.1 & 95.0$\pm$0.1 & 77.1$\pm$0.2 & 85.3$\pm$0.1 & 70.3$\pm$0.1 & 47.7$\pm$0.5 & 38.1$\pm$0.1 & 66.4 \\
            \textbf{MVRML} & 52.1$\pm$0.1 & 97.6$\pm$0.0 & 77.9$\pm$0.2 & 88.3$\pm$0.3 & 71.3$\pm$0.1 & 51.0$\pm$0.6 & 45.7$\pm$0.1 & 69.1 \\
            \bottomrule
        \end{tabular}
        \label{tab:domainbed} 
    \end{center}
    %\vspace{-15pt}
\end{table}
\textbf{Experiments on DomainBed.} We conduct an experiment on DomianBed benchmark, including CMNIST, RMNIST, VLCS, PACS, OfficeHome, TerraInc and DomainNet. The experiments are repeated three times with a learning rate of $1e-3$. Since our method consists of re-estimate Batch Normalization layer, the Batch Normalization layers are unfreezd. As shown in \cref{tab:domainbed}, ERM is the reported performance of DomainBed and ERM$^*$ is our reproduced performance with the same learning rate and freezed BN. Our method can achieve better performance than ERM (a strong baseline in DomainBed) on these datasets.

\begin{figure}[h]
    \centering
    \resizebox{.7\linewidth}{!}{
        \hfill
        \begin{subfigure}{0.49\linewidth}
            \includegraphics[width=1\linewidth]{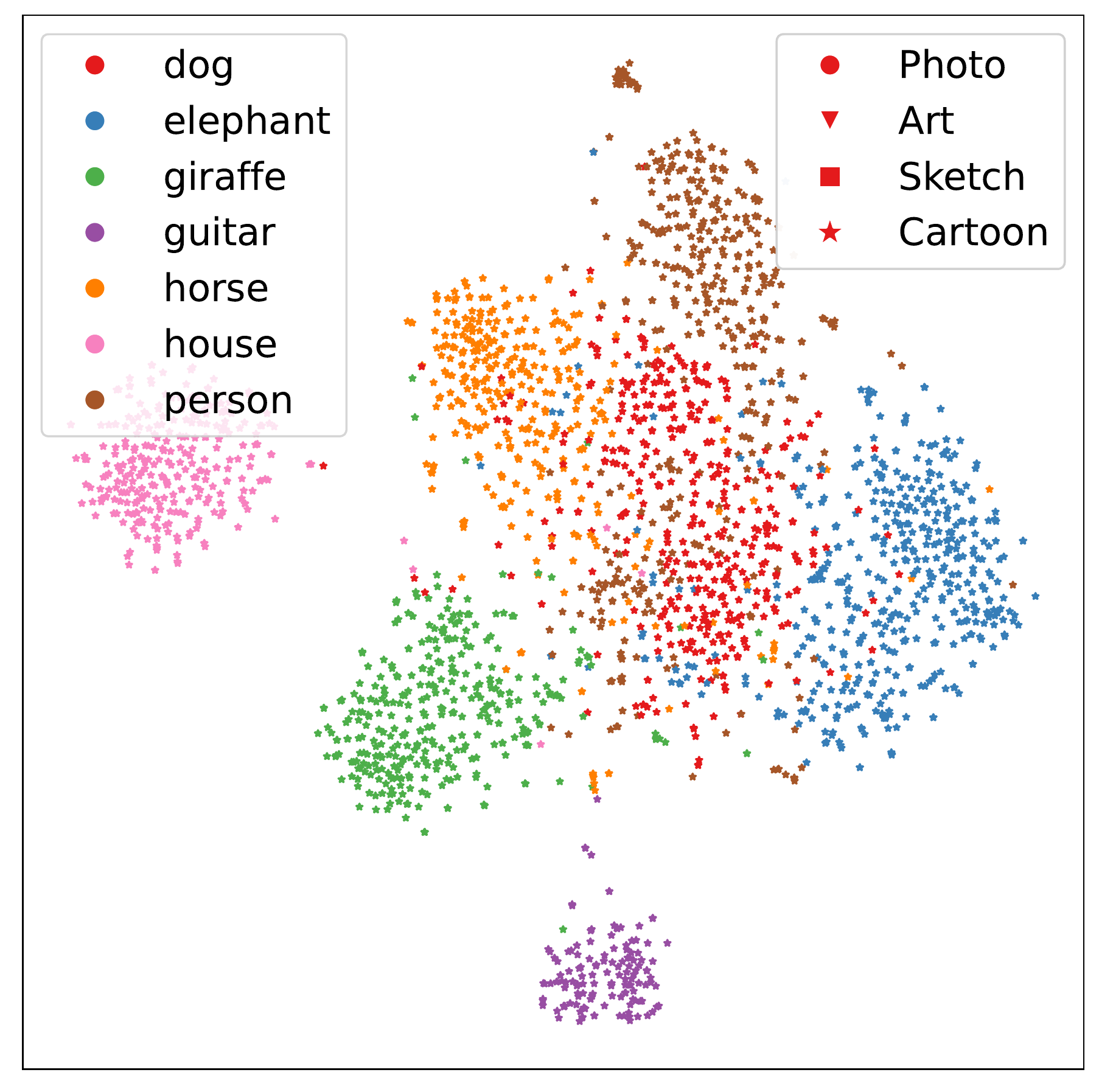}
            \caption{DeepAll}
        \end{subfigure}
        \hfill
        \hspace{50pt}
        \begin{subfigure}{0.49\linewidth}
            \includegraphics[width=1\linewidth]{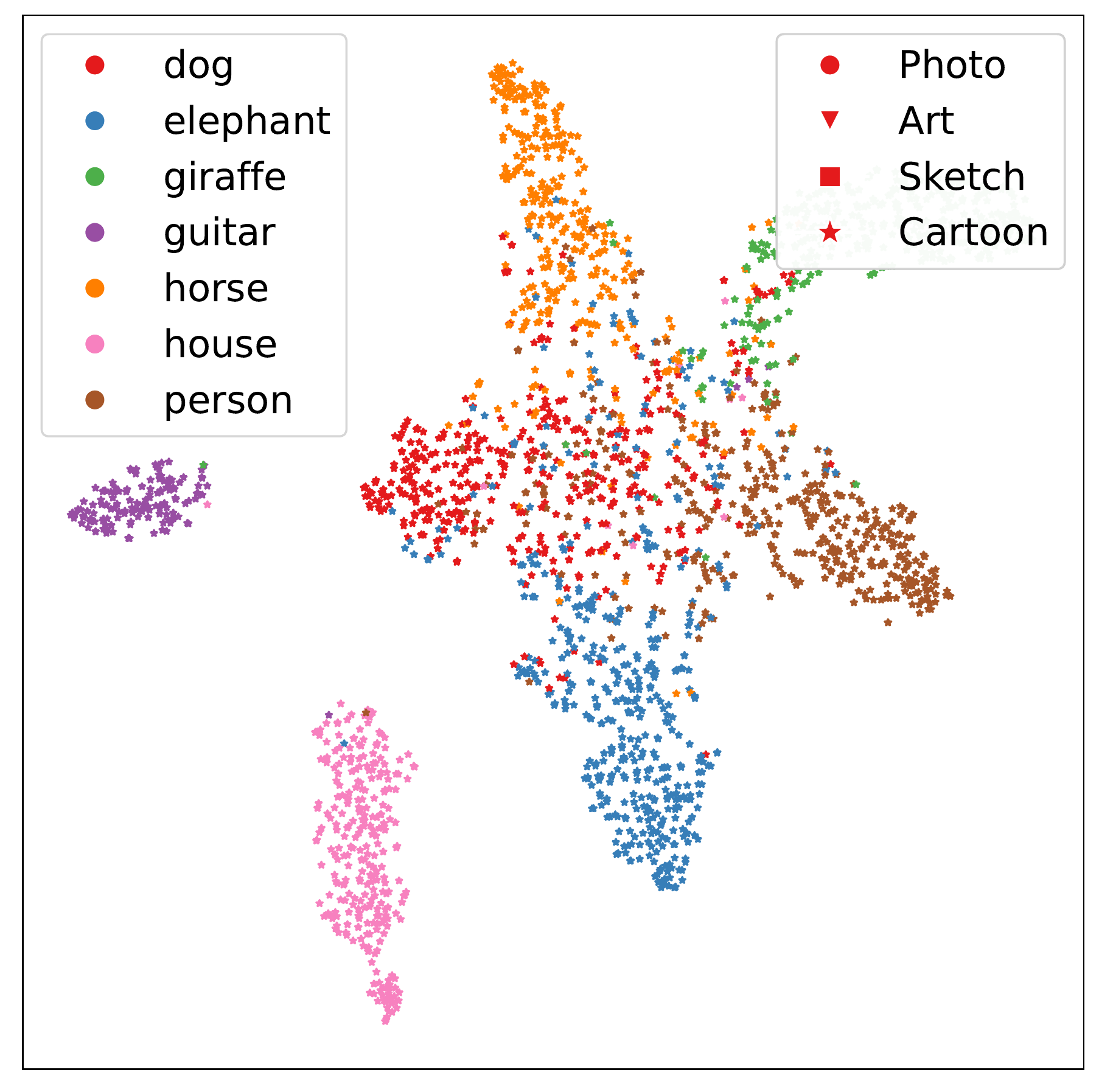}
            \caption{Ours}
        \end{subfigure}
        \hfill
    }
    \caption{The visualization of the feature learned on PACS dataset. DeepAll and the model trained with our method are shown in the figure. The target domain is Cartoon, and the others are all source domains. Different colors indicate different classes, and different shapes indicate different domains (best viewed in color).}
    \label{fig:tsne}
    %\vspace{-0.4cm}
\end{figure}

\textbf{Visualization of learned representation.} To qualitatively visualize the representation learned by our method, we generate the t-SNE map of both DeepAll and our model. The target domain is Cartoon, and we only utilize the validation set in the source domain and all images in the test domain to obtain the visualization result. The better the model can generalize, the more clustered the data should be. As shown in \cref{fig:tsne}, DeepAll cannot cluster the unseen samples well since the plain training cannot prevent overfitting. By contrast, MVRML can yield better clustering results, demonstrating its generalizability.

\section{Proof of Theorem 1}

\subsection{Notations}

{We denote the source domain as $\mathcal{S}=\{\mathcal{D}_1, ...\mathcal{D}_N\}$ and the target domain as $\mathcal{T}=\mathcal{D}_{N+1}$.} We denote a task as $t=(\mathcal{B}_{tr},\mathcal{B}_{te})$, which is obtained by sampling from $\mathcal{S}$. At each iteration, a sequence of sampled tasks along a single trajectory is defined as $\mathbf{T}=\{t_0, \dots, t_m\}$ with a size of $m$. Each task in $\mathbf{T}$ is sampled from a mixture distribution of source domains $\mathcal{U}=\sum_{\mathcal{D}_i\sim\mathcal{S}}\alpha_i\mathcal{D}_i$, where $\sum_i\alpha_i=1$. Each distribution $\mathcal{U}_j$ is sampled from a meta distribution $\mathcal{V}$ with uniformly distributed mixture cooefficients $\sum_i\alpha^j_i=1$. Thus, $\mathcal{V}$ and $\mathcal{S}$ are actually equivalent with respect to the data point $(x, y)$ since each data point appears with the same probability in $\mathcal{V}$ and $\mathcal{S}$.
The meta sequence for the meta-learning methods is defined as $\mathbb{T}=\{\mathbf{T}_0, \mathbf{T}_1, \dots, \mathbf{T}_n\}$ with a size of $n$.
A training algorithm $\mathbb{A}$ trained with $\mathbb{T}$ or $\mathbf{T}$ is denoted as $\theta=\mathbb{A}(\mathbb{T})$ or $\theta=\mathbb{A}(\mathbf{T})$. We define the expected risk as $\mathcal{E}_{\mathcal{P}}(\theta)=\mathbb{E}_{(x_j, y_j)\sim \mathcal{P}} \ell (f(x_j|\theta), y_j)$. With a little abuse of notation, we define the loss with respect to $\mathbf{T}$  as $\mathcal{L}(\mathbf{T};\theta)=\frac{1}{m}\sum_{(\mathcal{B}_{tr}, \mathcal{B}_{te})\in \mathbf{T}} \frac{1}{2}(\mathcal{L}(\mathcal{B}_{tr};\theta)+\mathcal{L}(\mathcal{B}_{te};\theta))$ and the loss with respect to $\mathbb{T}$  as $\mathcal{L}({\mathbb{T}};\theta)=\frac{1}{n}\sum_{\mathbf{T}\in \mathbb{T}} \mathcal{L}(\mathbf{T};\theta)$. \\

\subsection{Lemma}

\noindent \textbf{Lemma 1.} {Given two distributions $\mathcal{P}$ and $\mathcal{Q}$, the following inequality holds}~\cite{zhao2018adversarial}:
$$
    \mathcal{E}_{\mathcal{P}}(\theta) \le \mathcal{E}_{\mathcal{Q}}(\theta) + \frac{1}{2}\textbf{Div}(\mathcal{D}_{\mathcal{P}}, \mathcal{D}_{\mathcal{Q}}).
$$
where $\textbf{Div}$ is the divergence between two distributions.\\

{Since our training scheme is based on the meta-learning algorithm}, we introduce the generalization bound of meta-learning with respect to the sample size in \cite{al2021data}. By reformulating our training tasks mentioned above, we introduce the following lemma:

\noindent \textbf{Lemma 2.} {Assume that an algorithm $\mathbb{A}$ satisfies the following two conditions:}

\noindent {\textit{C1}. For every pair of meta sequences $\mathbb{T}=\{\mathbf{T}_0, ..., \mathbf{T}_n\}$, $\mathbb{T}^{\backslash i}:=\mathbb{T}\backslash \{\mathbf{T}_i\}$, and for every task sequence $\mathbf{T}$, we have $\left\|\mathcal{L}(\mathbf{T}; \mathbb{A}(\mathbb{T})) - \mathcal{L}(\mathbf{T}; \mathbb{A}(\mathbb{T}^{\backslash i}))\right\| \le \beta_1$.}

\noindent {\textit{C2}. For every pair of task sequences $\mathbf{T}=\{t^0, \dots, t^m\}$, $\mathbf{T}^{\backslash j}:=\mathbf{T}\backslash \{t_j\}$, and for any task $t$, we have $\left\|\mathcal{L}(t;\mathbb{A}(t)) - \mathcal{L}(t;\mathbb{A}(t^{\backslash j})), y)\right\| \le \beta_2$.}

\noindent {Then for any meta distribution $\mathcal{P}$, the following inequality holds} with probability at least $1-\delta$ \cite{al2021data}:
$$
    \mathcal{E}_{\mathcal{P}}(\theta)\le \hat{\mathcal{E}}_{\mathcal{P}}(\theta) + 2\beta_1 + (4n\beta_1+M)\sqrt{\frac{\ln \frac{1}{\delta}}{2n}} + 2\beta_2,
$$
where $M$ is a bound of loss function $\ell$. $ \hat{\mathcal{E}}_{\mathcal{P}}(\theta)$ is the empirical error on distribution $\mathcal{P}$.
%{$\beta_1$ and $\beta_2$ are functions of the number of task sequences $n$ and the number of tasks $m$ in each sequence.} 
When $\beta_1=o(1/n^a), a\ge1/2$ and $\beta_2=o(1/m^b),b\ge0$, this bound becomes non-trivial.

\textit{C1} and \textit{C2} are $\beta$-uniform stability conditions~\cite{bousquet2002stability} that indicate the sensitivity of the algorithm to the removal of an arbitrary point from the training sample, and if the uniform stability condition holds, we can upper bound the expected error by the empirical error.

\subsection{Proof}

\noindent \textbf{Theorem 1.} {Assume that algorithm $\mathbb{A}$ satisfies $\beta_1$-uniform stability}~\cite{bousquet2002stability} {with respect to $\mathcal{L}(\mathbb{T};\mathbf{A}(\mathbb{T}))$ and $\beta_2$-uniform stability with respect to $\mathcal{L}(\mathbf{T};\mathbf{A}(\mathbf{T}))$.  The following domain generalization error bound holds with probability at least $1-\delta$:
\begin{align*}
    \mathcal{E}_{\mathcal{T}}(\theta) \le \hat{\mathcal{E}}_{\mathcal{S}}(\theta) + \frac{1}{2}\sup_{\mathcal{D}_i \in \mathcal{S}}\textbf{Div}(\mathcal{D}_i, \mathcal{T}) + (2\beta_1 + (4n\beta_1+M)\sqrt{\frac{\ln \frac{1}{\delta}}{2n}} + 2\beta_2).
\end{align*}

\noindent \textit{Proof}.
Consider a mixture distribution of $N$ source domains where the mixture weight is geven by $\gamma$ and $\sum_{i=1}^N\gamma_i=1$.
\begin{flalign}
    \mathcal{E}_{\mathcal{T}}(\theta) & \le \mathcal{E}_{S}(\theta) +  \frac{1}{2}\textbf{Div}(\mathcal{S}, \mathcal{T})  \le \mathcal{E}_{S}  (\theta) +  \frac{1}{2}\sum_i^N \gamma_i\textbf{Div}(\mathcal{D}_{i}, \mathcal{T})                        \\
                                      & \le \mathcal{E}_{S}(\theta) + \frac{1}{2}\sup_{\mathcal{D}_i \in \mathcal{S}}\textbf{Div}(\mathcal{D}_i, \mathcal{T})                                                                                            \\
                                      & \le \hat{\mathcal{E}}_{\mathcal{S}}(\theta) + \frac{1}{2}\sup_{\mathcal{D}_i \in\mathcal{S}}\textbf{Div}(\mathcal{D}_i, \mathcal{T}) + 2\beta_1 + (4n\beta_1+M)\sqrt{\frac{\ln \frac{1}{\delta}}{2n}} + 2\beta_2
\end{flalign}

We first bound the expected error between the source and target domains from Eq. (1) to Eq. (2) with \textbf{Lemma 1}. Eq. (1) is obtained according to \cite{zhao2018adversarial} that  $\textbf{Div}(\mathcal{S}, \mathcal{T}) \le \sum_i^N \gamma_i\textbf{Div}(\mathcal{D}_{i}, \mathcal{T})$, where $\sum_{i=1}^N\gamma_i=1$. {Then we bound Eq. (1) with the maximum divergence  $\sup_{\mathcal{D}_i \in \mathcal{S}}\textbf{Div}(\mathcal{D}_i, \mathcal{T})$ between $\mathcal{D}_i$ and $\mathcal{T}$.}

According to C.1 and C.2 and the fact that $\mathcal{E}_{\mathcal{V}}(\theta)=\mathcal{E}_{\mathcal{S}}(\theta)$ since the distribution of $\mathcal{V}$ and $\mathcal{S}$ are equivalent with respect to the data points, we have:
\begin{flalign}
    \mathcal{E}_{\mathcal{V}}(\theta)\le \hat{\mathcal{E}}_{\mathcal{V}}(\theta) + 2\beta_1 + (4n\beta_1+M)\sqrt{\frac{\ln \frac{1}{\delta}}{2n}} + 2\beta_2.
\end{flalign}
{Thus, by replacing the expected error of source domains in Eq. (3) with empirical error in Eq. (4), we arrive at Theorem 1.}

\section{Other Details}

\textbf{Visualization of loss surface.} To visualize the loss surface of a model, we follow the visualization technique in \cite{garipov2018loss}. Suppose we have weight vectors of three models $w_1, w_2, w_3$. We first find two basis $\hat{u}, \hat{v}$ of the plane across these weights: $u=\left(w_{2}-w_{1}\right)$, $v=({\left(\theta_{3}-\theta_{1}\right)-\left\langle\theta_{3}-\theta_{1}, \theta_{2}-\theta_{1}\right\rangle})/\left\|\theta_{2}-\theta_{1}\right\|^{2}\cdot\left(\theta_{2}-\theta_{1}\right)$,
$\hat{u}=u /\|u\|, \hat{v}=v /\|v\|$. Then, we deﬁne a Cartesian grid in the basis $\hat{u}, \hat{v}$ and calculate the training/test loss corresponding to each of the points in the grid.

\end{document}